\documentclass[journal]{IEEEtran}
\usepackage{amsmath,amsfonts,amssymb}
\usepackage{algorithmic}
\usepackage{algorithm}
\usepackage{array}
\usepackage[caption=false,font=normalsize,labelfont=sf,textfont=sf]{subfig}
\usepackage{textcomp}
\usepackage{stfloats}
\usepackage{url}
\usepackage{graphicx}
\usepackage{cite}
\usepackage[table]{xcolor}
\usepackage{booktabs}
\usepackage{multirow}
\usepackage{adjustbox}
\usepackage{pifont}
\usepackage{balance}
\newcommand{\cmark}{\ding{51}}%
\newcommand{\xmark}{\ding{55}}%
\graphicspath{{figures/}}
\newcommand{\tblvenue}[1]{{\color{gray!55}\textit{#1}}}
\newcommand{\tblmethod}[3]{#1~\cite{#2}\ \tblvenue{#3}}
\newcommand{\tblencoder}[3]{#1~\cite{#2}\ \tblvenue{#3}}

\begin{document}

\renewcommand{\IEEEtitletopspaceextra}{-36pt}
\title{%
{\normalfont\footnotesize\itshape
Submitted to IEEE Transactions on Image Processing on 12 July 2026; currently under review. Version 1 is identical to the original journal submission. Version 2 differs from Version 1 only by the addition of this submission-history note; the manuscript content is otherwise unchanged.\par}
\vspace{0.25em}
Complementarity-Driven Dual-Teacher Hierarchical Alignment for Zero-Shot EEG-to-Image Retrieval}
\author{Lukun Wu, Jie Li, Ziqi Ren, Ruiyang Xia, Yan Gao, and Xinbo Gao,~\IEEEmembership{Fellow,~IEEE,}%
\thanks{Lukun Wu, Jie Li, and Xinbo Gao are with the Visual Information Processing Laboratory, School of Electronic Engineering, Xidian University, Xi'an 710071, China (Corresponding authors: Jie Li; Xinbo Gao, e-mail: leejie@mail.xidian.edu.cn; xbgao@mail.xidian.edu.cn).}%
\thanks{Ziqi Ren is with the Center for Biomedical-photonics and Molecular Imaging (BMI), School of Life Science and Technology, Xidian University, Xi'an 710126, China.}

\thanks{Ruiyang Xia is with the School of Computer Science and Technology, Chongqing University of Posts and Telecommunications, Chongqing 400065, China.}
\thanks{Yan Gao is with the School of Electronic Information (School of Artificial Intelligence), Northwest University, Xi'an 710127, China.}}
\markboth{}%
{Wu \MakeLowercase{\textit{et al.}}: Complementarity-Driven DTHA for Zero-Shot EEG-to-Image Retrieval}

\maketitle

\begin{abstract}
  Decoding visual information from EEG is central to brain--computer interfaces and brain-inspired intelligence, and contrastive alignment with pretrained vision foundation models (VFMs) has become the dominant paradigm.
  However, existing methods align EEG solely with CLIP-family VFMs, leaving hierarchical visual information in neural responses under-exploited.
  To address this gap, we formulate a mechanism--paradigm correspondence view that links hierarchical visual processing (from appearance to semantics) to the complementary biases of cross-paradigm VFMs, with image-only models (e.g., DINO) and image--text models (e.g., CLIP) capturing distinct facets of this hierarchy.
  Guided by this view, we propose the Dual-Teacher Hierarchical Alignment (DTHA) framework.
  Concretely, a shared EEG encoder is supervised by frozen multi-blur CLIP and shallow DINOv2 teachers via dual-branch contrastive alignment, so that each teacher branch can fully supervise the visual cues it emphasizes.
  A fused-retrieval objective learns adaptive weights to combine their complementary visual representations for zero-shot EEG-to-image retrieval.
  For the shared EEG encoder, we introduce the Attentive Modulated Scalp-Pattern Bank Encoder (AMSPBE), which combines temporo-spatial attention with a learnable scalp-pattern bank for accurate, interpretable visual feature extraction, while further modeling regional temporal dynamics and filter-bank topographies.
  On THINGS-EEG zero-shot retrieval, DTHA achieves 92.4\% Top-1 accuracy, exceeding prior state of the art by 12.4 percentage points.
  Spatial, spectral, and representational-similarity analyses, together with THINGS-MEG evaluation, collectively validate DTHA's effectiveness, biological plausibility, and cross-modality generalization.
  This complementarity-driven perspective shows that supervision from cross-paradigm VFMs enables more complete extraction of hierarchical visual information from brain activity.
  It further highlights cross-paradigm brain--image alignment as a promising direction for future work.
\end{abstract}


\begin{figure}[!t]
  \centering
  \includegraphics[width=\columnwidth]{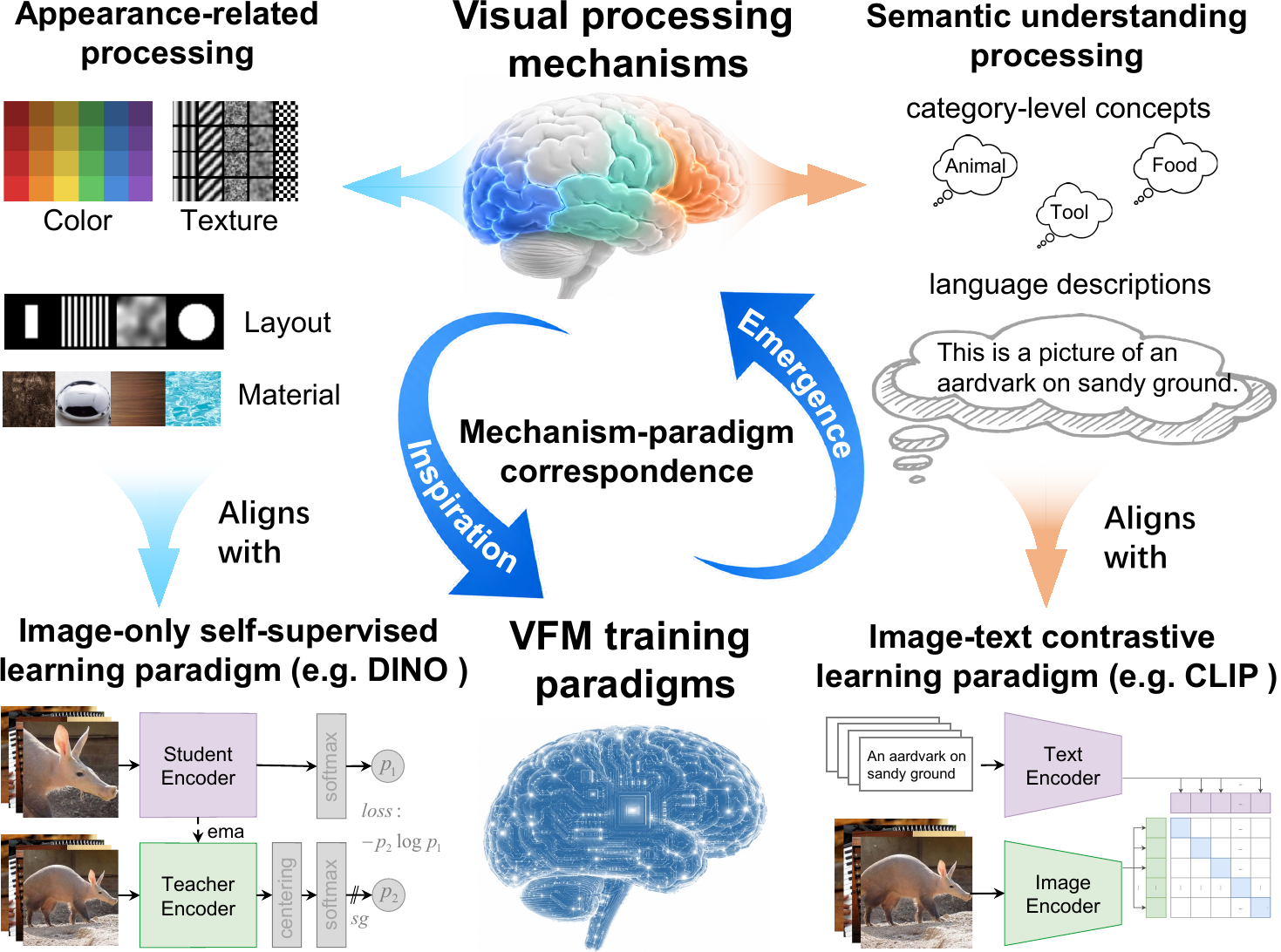}
  \caption{Mechanism--paradigm correspondence between hierarchical visual processing (appearance-related structure to semantic understanding) and emergent VFM representations from distinct training paradigms (image-only self-supervised vs.\ image--text contrastive learning), motivating complementarity-driven dual-teacher EEG alignment beyond CLIP-family VFMs alone.}
  \label{fig:motivation}
\end{figure}

\section{Introduction}
\label{sec:introduction}

\IEEEPARstart{V}{isual} neural decoding aims to extract visual information from brain activity, serving as a bridge between human cognition and artificial intelligence while deepening our understanding of the human visual mechanism and advancing brain-inspired intelligence.
Among non-invasive neuroimaging modalities, electroencephalography (EEG) offers high temporal resolution and portability, and has drawn increasing attention for brain--computer interface (BCI) applications~\cite{spampinato2017deep, gifford2022}.
However, the complexity of human visual processing and the relatively low signal-to-noise ratio of scalp EEG pose significant challenges for EEG-based visual decoding.
Recent large-scale and diverse datasets acquired under rapid serial visual presentation (RSVP) paradigms have provided a strong empirical foundation for this research line~\cite{gifford2022, hebart2023}.
On these benchmarks, zero-shot EEG-to-image retrieval has become a standard evaluation task for visual decoding, with retrieval accuracy reflecting decoding performance from scalp EEG.

The prevailing paradigm for EEG-based visual decoding is contrastive learning, which aligns EEG with pretrained vision foundation models (VFMs) in a shared embedding space to extract visual information from EEG~\cite{song2024decoding,li2024visual,chen2024visual,wei2024mb2c}.
This paradigm has driven rapid progress, yet most methods under it align EEG only with CLIP-family VFMs.
CLIP supervision recovers category-level semantics effectively, but under-extracts appearance-related structure that is also encoded in EEG.

Decades of neuroscience research have established that human vision is organized hierarchically rather than as a single representational stage~\cite{dicarlo2012how,kravitz2013}.
As illustrated in Fig.~\ref{fig:motivation} (top), this hierarchy extends from appearance-related structure toward semantic understanding of visual content~\cite{patterson2007}.
In computer vision, modern VFMs are trained under distinct paradigms (Fig.~\ref{fig:motivation}, bottom).
\textbf{Image-only self-supervised learning} (e.g., DINO) learns structure from visual data without language supervision~\cite{caron2021,oquab2024}, whereas \textbf{image--text contrastive learning} (e.g., CLIP) aligns visual embeddings with natural-language descriptions~\cite{radford2021}.
Driven by scale in data and model capacity, modern VFMs exhibit an \textbf{emergence} of paradigm-differentiated visual capabilities that parallel human vision: for example, DINO develops dense patch-level descriptors that preserve appearance-related structure, whereas CLIP learns transferable image--text representations that support semantic categorization.
Recent brain--model studies of DINOv3 further suggest that its representations can functionally relate to the human visual mechanism~\cite{raugel2026}.
Integrating these observations, we formulate a \textbf{mechanism--paradigm correspondence} view that links hierarchical visual processing to VFMs trained under distinct paradigms for brain--image alignment.

Guided by this view, we propose \textbf{DTHA} (Dual-Teacher Hierarchical Alignment), a complementarity-driven dual-teacher alignment framework.
Concretely, a shared encoder learns an EEG embedding that is contrastively aligned to frozen multi-blur CLIP and shallow DINOv2 teachers through dual-branch supervision, so that each teacher can fully supervise the visual cues it emphasizes.
Complementing these branch objectives, a fused-retrieval objective learns adaptive fusion weights during training to merge branch-wise brain--image similarities into a joint ranking score, so that semantic and appearance-related cues jointly drive zero-shot EEG-to-image retrieval.
We further introduce \textbf{AMSPBE} (Attentive Modulated Scalp-Pattern Bank Encoder), which applies temporo-spatial attention modulation and a learnable scalp-pattern bank for more accurate and interpretable EEG feature extraction, including regional temporal dynamics and filter-bank topographies.
We also conduct spatial, spectral, and representational similarity analysis (RSA) to assess the biological plausibility of our dual-teacher framework.

Our main contributions are summarized as follows:
\begin{enumerate}
\item To our knowledge, we are the first to formulate a mechanism--paradigm correspondence view between hierarchical human visual processing and cross-paradigm VFMs, and to apply it toward more complete extraction of visual information from EEG.
\item We propose DTHA (Dual-Teacher Hierarchical Alignment), a complementarity-driven dual-teacher framework that enables full exploitation of each teacher branch and effective fusion of their complementary visual cues. We further propose AMSPBE (Attentive Modulated Scalp-Pattern Bank Encoder) for accurate and interpretable extraction of visual features from EEG.
\item Through extensive experiments and analyses, we validate the rationale, effectiveness, and cross-modality generalization of our approach, achieving state-of-the-art zero-shot retrieval on standard EEG and MEG benchmarks. Spatial, spectral, and representational analyses further support the biological plausibility of complementarity-driven dual-teacher alignment and provide new insights into visual neural decoding.
\end{enumerate}

\begin{figure*}[!t]
  \centering
  \includegraphics[width=\textwidth]{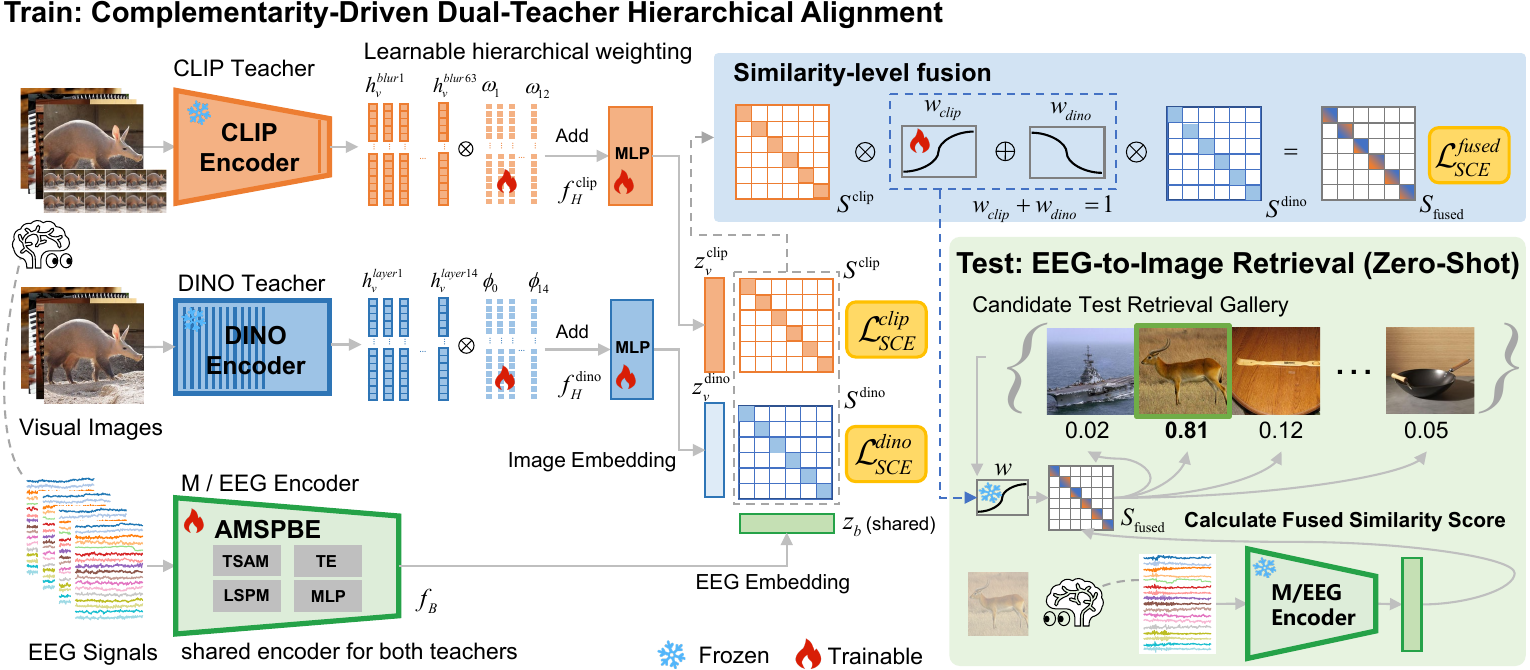}
  \caption{Overview of complementarity-driven dual-teacher hierarchical alignment (DTHA) for zero-shot EEG-to-image retrieval.
  \textbf{Train:} frozen multi-blur CLIP-RN50 and shallow DINOv2 ViT-L/14 teachers supply $z_v^{\mathrm{clip}}$ and $z_v^{\mathrm{dino}}$ through learnable hierarchical weighting modules $f_H^{\mathrm{clip}}$ and $f_H^{\mathrm{dino}}$; shared AMSPBE ($f_B$) maps RSVP-evoked EEG to brain embedding $z_b$, which is aligned to both teachers via branch-wise SCE losses $\mathcal{L}^{\mathrm{clip}}_{\mathrm{SCE}}$ and $\mathcal{L}^{\mathrm{dino}}_{\mathrm{SCE}}$.
  Fusion weights $w_{\mathrm{clip}}$ and $w_{\mathrm{dino}}$ combine similarity matrices $S^{\mathrm{clip}}$ and $S^{\mathrm{dino}}$ into $S_{\mathrm{fused}}$ for fused-retrieval learning $\mathcal{L}^{\mathrm{fused}}_{\mathrm{SCE}}$ (similarity-level fusion, not feature concatenation).
  \textbf{Test:} frozen $f_B$, $f_H^{\mathrm{clip}}$, and $f_H^{\mathrm{dino}}$ encode query EEG and gallery images; learned $w_{\mathrm{clip}}$ and $w_{\mathrm{dino}}$ fuse $S^{\mathrm{clip}}$ and $S^{\mathrm{dino}}$ for gallery ranking.}
  \label{fig:method}
\end{figure*}

\section{Related Work}
\label{sec:related_work}

\subsection{EEG Visual Decoding}
\label{sec:rw_eeg_decoding}

Visual neural decoding aims to interpret brain activity to retrieve, recognize, or reconstruct visual stimuli.
While early efforts often relied on fMRI for its high spatial resolution~\cite{ren2021reconstructing,takagi2023high,scotti2024reconstructing,li2025neuraldiffuser,chen2025mindgpt}, electroencephalography (EEG) has gained increasing attention for its high temporal resolution, non-invasive acquisition, and portability in brain--computer interface (BCI) applications.
Following early deep-learning evidence that object-evoked EEG can support visual classification~\cite{spampinato2017deep}, large-scale RSVP benchmarks such as THINGS-EEG~\cite{gifford2022} and frameworks like BraVL~\cite{du2023decoding} and NICE~\cite{song2024decoding} established zero-shot EEG-to-image retrieval as a central research setting.
Contrastive alignment with frozen CLIP-family VFMs is now the dominant paradigm on these benchmarks.

Recent methods improve decoding along several directions, yet most still assemble supervision from CLIP-family or generative-visual teachers.
\textbf{Student-side encoding} improves EEG representation learning, e.g., subject-aggregation and group-aware variants of NICE (NICE-SA/GA)~\cite{song2024decoding} and ATM with a patch-based spatiotemporal-convolution encoder~\cite{li2024visual}.
\textbf{Alignment constraint reinforcement} strengthens brain--image alignment with dedicated objectives, including VE-SDN~\cite{chen2024visual} and MB2C~\cite{wei2024mb2c}.
\textbf{Information enhancement} incorporates auxiliary modalities, such as language-guided contrastive learning in NICE++~\cite{song2025recognizing} and depth-aware supervision in CogCap~\cite{zhang2025cognitioncapturer} and CogCapPro~\cite{zhang2026cognitioncapturerpro}; in each case, language descriptions or depth maps are still CLIP-family encoded for alignment.
\textbf{Teacher-side adaptation} modifies how frozen CLIP features are presented to the brain branch, e.g., uncertainty-aware blur priors in UBP~\cite{wu2025ubp}, multi-blur CLIP fusion in LVBP~\cite{liu2026visualeeg}, visual prompt tuning in NeuroCLIP~\cite{wang2025neuroclip}, cognitive-prior augmentation in NeuroBridge~\cite{zhang2026neurobridge}, multi-encoder fusion of CLIP and VAE features in HVF~\cite{zheng2026hvf}, and adaptive CLIP feature shrinking in ATS (our prior work)~\cite{wu2026ats}.
Although these advances have substantially improved retrieval performance, the underlying teacher-side supervision still favors category-level semantics and only limited appearance-related cues---whether from multi-blur CLIP inputs or VAE latents---and rarely employs cross-paradigm teacher pairing, leaving richer fine-grained structure in EEG under-exploited.
This gap motivates our mechanism--paradigm correspondence view and complementarity-driven cross-paradigm dual-teacher alignment (DTHA).

\subsection{Vision Foundation Models}
\label{sec:rw_vfm}

Visual foundation models (VFMs) are large-scale encoders pretrained on broad visual or multimodal data and reused across downstream vision tasks.
Although modern VFMs increasingly share transformer backbones trained on web-scale data, their behavior differs less by architectural family alone than by \emph{training paradigm}.

We adopt a coarse two-way categorization of VFMs: \emph{image-only} supervision and \emph{image--text} alignment.
Within the image-only line, multi-view self-distillation (DINO, DINOv2, DINOv3~\cite{simeoni2025dinov3}) learns invariances across augmented views and yields dense patch-level descriptors that emphasize appearance-related structure.
The image--text line aligns visual embeddings with language supervision (CLIP, PE-Core~\cite{bolya2025pecore}), favoring transferable semantic categorization.
As comparative analyses confirm~\cite{walmer2023}, distinct training paradigms systematically shape attention patterns, feature geometry, and downstream representational biases.
Aligning EEG to frozen MoCo-, MAE-, or SAM-pretrained VFMs remains promising but is outside the scope of this work.

\subsection{Brain--Model Comparison Studies}
\label{sec:rw_brain_model}

Brain--model comparison studies, using encoding models and RSA among other methods, analyze the brain--model mapping between human visual responses and deep network representations~\cite{kriegeskorte2008,yamins2014,gucclu2015,conwell2024}.
More recently, Raugel et al.~\cite{raugel2026} analyze DINOv3 in depth, examining how model scale, training duration, and data type shape alignment with human fMRI/MEG.
These studies are highly inspiring, yet they primarily analyze brain--model relationships rather than apply such insights to visual decoding, and the recent DINOv3 analysis remains within a single self-supervised paradigm.
In contrast, we formulate a \textbf{mechanism--paradigm correspondence} view that exploits complementary \textbf{cross-paradigm} VFMs for improved EEG visual decoding.

\section{Methodology}
\label{sec:methodology}

\subsection{Problem Formulation}
\label{sec:formulation}
We study zero-shot EEG-to-image retrieval on RSVP-evoked recordings following the standard NICE setup~\cite{song2024decoding}.
Given training data $D_{\mathrm{train}} = \{ (x_v, x_b, y) \}$, where $x_v$ is an image, $x_b$ the paired brain signal, and $y \in Y_{\mathrm{seen}}$ is the class label, the model is trained only on seen classes.
As illustrated in Fig.~\ref{fig:method}, at test time the model is evaluated on $D_{\mathrm{test}} = \{ (x_v^u, x_b^u, y^u) \}$, where $y^u \in Y_{\mathrm{unseen}}$ and $Y_{\mathrm{seen}} \cap Y_{\mathrm{unseen}} = \emptyset$.
For a query brain signal $x_b^u$, the goal is to retrieve its corresponding image $x_v^u$ from gallery $X_v^u$ (all images from $Y_{\mathrm{unseen}}$) by ranking candidates according to brain--image similarity in the learned latent space.

\subsection{DTHA: Dual-Teacher Hierarchical Alignment}
\label{sec:dtha}

Fig.~\ref{fig:method} illustrates \textbf{DTHA}, a complementarity-driven dual-teacher framework.
During training, frozen pretrained CLIP-RN50 extracts final-layer features from $M{=}12$ blur-scaled inputs, and frozen pretrained DINOv2 ViT-L/14 extracts shallow-layer CLS features at blocks $0$--$14$; $f_H^{\mathrm{clip}}$ and $f_H^{\mathrm{dino}}$ (Sec.~\ref{sec:teacher_fusion}) weight and project these features to teacher embeddings $z_v^{\mathrm{clip}}$ and $z_v^{\mathrm{dino}}$.
A shared AMSPBE encoder $f_B$ maps EEG to $z_b \in \mathbb{R}^{1024}$, which is contrastively aligned with $z_v^{\mathrm{clip}}$ and $z_v^{\mathrm{dino}}$ in the CLIP and DINO branches, respectively; $f_H^{\mathrm{clip}}$, $f_H^{\mathrm{dino}}$, $f_B$, and the fusion logits are trainable.

The alignment between each teacher embedding and the student is optimized with a symmetric contrastive objective.
Within a mini-batch of size $N$, each paired image--brain sample is a positive pair and all other cross-pairings are negatives.
Let $z_{b,i}$ and $z_{v,i}$ denote the brain and teacher embeddings of the $i$-th sample.
We adopt the Symmetric Cross-Entropy (SCE) loss, an objective derived from InfoNCE~\cite{radford2021,oord2019}:
\begin{multline}
\label{eq:sce}
\mathcal{L}_{\mathrm{SCE}}
= -\frac{1}{2N} \sum_{i=1}^{N} \Biggl[ \log \frac{\exp(z_{b,i}^\top z_{v,i}/\tau)}{\sum_{k=1}^{N} \exp(z_{b,i}^\top z_{v,k}/\tau)} \\
+ \log \frac{\exp(z_{v,i}^\top z_{b,i}/\tau)}{\sum_{k=1}^{N} \exp(z_{v,i}^\top z_{b,k}/\tau)} \Biggr],
\end{multline}
where $(\cdot)^\top(\cdot)$ denotes the inner product and $\tau{=}1$ is fixed in all reported experiments.
Substituting $z_{v,i}^{\mathrm{clip}}$ for $z_{v,i}$ in Eq.~\eqref{eq:sce} yields $\mathcal{L}^{\mathrm{clip}}_{\mathrm{SCE}}$; substituting $z_{v,i}^{\mathrm{dino}}$ likewise yields $\mathcal{L}^{\mathrm{dino}}_{\mathrm{SCE}}$:
\begin{equation}
\begin{aligned}
\mathcal{L}^{\mathrm{clip}}_{\mathrm{SCE}}
&\triangleq \mathcal{L}_{\mathrm{SCE}}\ \text{with}\ z_{v,i}\leftarrow z_{v,i}^{\mathrm{clip}}, \\
\mathcal{L}^{\mathrm{dino}}_{\mathrm{SCE}}
&\triangleq \mathcal{L}_{\mathrm{SCE}}\ \text{with}\ z_{v,i}\leftarrow z_{v,i}^{\mathrm{dino}}.
\end{aligned}
\label{eq:branch_loss}
\end{equation}

Under $\mathcal{L}^{\mathrm{clip}}_{\mathrm{SCE}}$ and $\mathcal{L}^{\mathrm{dino}}_{\mathrm{SCE}}$, $z_b$ is encouraged to capture the semantic and appearance-related cues emphasized by each teacher, respectively.
We further introduce a fused-retrieval objective that \textbf{learns similarity-level fusion weights} to combine branch similarities during training and \textbf{applies the same fusion rule at inference} (Fig.~\ref{fig:method}).
Stacking mini-batch embeddings as $Z_b=[z_{b,1},\ldots,z_{b,N}]^\top$ and $Z_v^{\mathrm{clip}}=[z_{v,1}^{\mathrm{clip}},\ldots,z_{v,N}^{\mathrm{clip}}]^\top$ (likewise $Z_v^{\mathrm{dino}}$), we form branch similarity matrices
\begin{equation}
\begin{aligned}
S^{\mathrm{clip}} &= Z_b(\bar{Z}_v^{\mathrm{clip}})^\top, \\
S^{\mathrm{dino}} &= Z_b(\bar{Z}_v^{\mathrm{dino}})^\top,
\end{aligned}
\label{eq:sim_matrices}
\end{equation}
where $\bar{Z}_v$ denotes the row-wise $\ell_2$-normalized teacher embeddings in the mini-batch.
To learn how much each branch should contribute to fused retrieval, trainable fusion logits $\eta_{\mathrm{clip}}$ and $\eta_{\mathrm{dino}}$ define fusion weights
\begin{equation}
\left[w_{\mathrm{clip}}, w_{\mathrm{dino}}\right]^{\top}
= \mathrm{softmax}\!\left(\left[\eta_{\mathrm{clip}}, \eta_{\mathrm{dino}}\right]\right),
\label{eq:fusion_weights}
\end{equation}
and the fused similarity matrix is
\begin{equation}
S_{\mathrm{fused}} = w_{\mathrm{clip}}\,S^{\mathrm{clip}} + w_{\mathrm{dino}}\,S^{\mathrm{dino}}.
\label{eq:fused_sim}
\end{equation}
Minimizing $\mathcal{L}^{\mathrm{fused}}_{\mathrm{SCE}}$ on $S_{\mathrm{fused}}$ drives $w_{\mathrm{clip}}$ and $w_{\mathrm{dino}}$ toward a branch combination that improves fused ranking accuracy.
The fused-retrieval objective applies the same SCE form to $S_{\mathrm{fused}}$:
\begin{multline}
\label{eq:fused_sce}
\mathcal{L}^{\mathrm{fused}}_{\mathrm{SCE}}
= -\frac{1}{2N} \sum_{i=1}^{N} \Biggl[ \log \frac{\exp(S_{\mathrm{fused},ii})}{\sum_{k=1}^{N} \exp(S_{\mathrm{fused},ik})} \\
+ \log \frac{\exp(S_{\mathrm{fused},ii})}{\sum_{k=1}^{N} \exp(S_{\mathrm{fused},ki})} \Biggr].
\end{multline}
The overall training objective is
\begin{equation}
\mathcal{L}
= \alpha\,\mathcal{L}^{\mathrm{clip}}_{\mathrm{SCE}}
+ \beta\,\mathcal{L}^{\mathrm{dino}}_{\mathrm{SCE}}
+ \gamma\,\mathcal{L}^{\mathrm{fused}}_{\mathrm{SCE}},
\label{eq:total_loss}
\end{equation}
with loss weights $(\alpha,\beta,\gamma){=}(1,1,0.25)$ by default.
At test time, each gallery image and query EEG are encoded by the frozen $f_H^{\mathrm{clip}}$, $f_H^{\mathrm{dino}}$, and $f_B$ to form branch similarity matrices $S^{\mathrm{clip}}$ and $S^{\mathrm{dino}}$ (Eq.~\eqref{eq:sim_matrices}); gallery images are ranked by $S_{\mathrm{fused}}$ using the fusion weights $w_{\mathrm{clip}}$ and $w_{\mathrm{dino}}$ learned during training (Eq.~\eqref{eq:fused_sim}).
Algorithm~\ref{alg:dtha} summarizes the training and inference procedure.

\begin{algorithm}[!t]
  \caption{Dual-Teacher Hierarchical Alignment (DTHA)}
  \label{alg:dtha}
  \begin{algorithmic}[1]
    \REQUIRE Multimodal training dataset $D_{\mathrm{train}}$; frozen CLIP/DINO backbones
    \ENSURE Trained $f_H^{\mathrm{clip}}$, $f_H^{\mathrm{dino}}$, $f_B$, and fusion logits $\{\eta_{\mathrm{clip}}, \eta_{\mathrm{dino}}\}$
    \STATE \textbf{Initialize} $f_H^{\mathrm{clip}}$ ($\theta^{\mathrm{clip}}$), $f_H^{\mathrm{dino}}$ ($\theta^{\mathrm{dino}}$), brain encoder $f_B$ (AMSPBE, $\phi$), and $\{\eta_{\mathrm{clip}}, \eta_{\mathrm{dino}}\}$
    \FOR{each iteration}
        \STATE Sample a mini-batch $\{(x_v, x_b)\}$ from $D_{\mathrm{train}}$
        \STATE Build teacher embeddings $z_v^{\mathrm{clip}}$ and $z_v^{\mathrm{dino}}$ (Sec.~\ref{sec:teacher_fusion}); compute $z_b \gets f_B(x_b)$
        \STATE Compute $\mathcal{L}^{\mathrm{clip}}_{\mathrm{SCE}}$, $\mathcal{L}^{\mathrm{dino}}_{\mathrm{SCE}}$, and $\mathcal{L}^{\mathrm{fused}}_{\mathrm{SCE}}$ (Eqs.~\eqref{eq:sce}--\eqref{eq:fused_sce})
        \STATE $\mathcal{L} \gets \alpha \mathcal{L}^{\mathrm{clip}}_{\mathrm{SCE}} + \beta \mathcal{L}^{\mathrm{dino}}_{\mathrm{SCE}} + \gamma \mathcal{L}^{\mathrm{fused}}_{\mathrm{SCE}}$; update $\theta^{\mathrm{clip}}$, $\theta^{\mathrm{dino}}$, $\phi$, and fusion logits by gradient descent
    \ENDFOR
    \STATE \textbf{Inference:} encode query $x_b$ and each gallery image $x_v$ with frozen $f_B$, $f_H^{\mathrm{clip}}$, and $f_H^{\mathrm{dino}}$
    \STATE Form $S^{\mathrm{clip}}$ and $S^{\mathrm{dino}}$, combine them into $S_{\mathrm{fused}}$ with the learned $w_{\mathrm{clip}}$ and $w_{\mathrm{dino}}$ (Eqs.~\eqref{eq:sim_matrices}--\eqref{eq:fused_sim}), and rank gallery images
  \end{algorithmic}
\end{algorithm}

\subsection{Learnable Hierarchical Weighting on the Image Side}
\label{sec:teacher_fusion}

Because different blur scales and network depths carry distinct visual cues, fixed pooling over CLIP blurs or DINO layers is suboptimal for brain--image alignment; we therefore learn softmax-normalized weights over these frozen hierarchical features when forming $z_v^{\mathrm{clip}}$ and $z_v^{\mathrm{dino}}$ (representational motivation in Sec.~\ref{sec:teacher_repr_diff}).

\paragraph{Frozen hierarchical features.}
For CLIP, let $\mathcal{B}_m(\cdot)$ denote Gaussian blur with scale $\sigma_m$.
We extract $M{=}12$ CLIP image embeddings
\begin{equation}
h_v^{(m)} = f_{\mathrm{CLIP}}\!\left(\mathcal{B}_m(x_v)\right), \quad m=1,\ldots,M,
\label{eq:clip_raw}
\end{equation}
with $\sigma \in \{1,3,9,\ldots,63\}$, as in LVBP~\cite{liu2026visualeeg}.
Blurred images weaken semantics while strengthening lower-level cues, thereby supplementing CLIP.
For DINOv2, $h_v^{(\ell)} = f_{\mathrm{DINO}}^{(\ell)}(x_v)$ denotes the $\ell_2$-normalized CLS token after block $\ell$; we use shallow layers $\ell \in \{0,\ldots,14\}$ because they better preserve appearance-related structure (Sec.~\ref{sec:teacher_repr_diff}).

\paragraph{Learnable hierarchical weighting.}
We use learnable hierarchical weighting modules $f_H^{\mathrm{clip}}$ and $f_H^{\mathrm{dino}}$ to map extracted hierarchical features to teacher embeddings via soft hierarchical weights followed by a lightweight MLP.
Given $H_{\mathrm{clip}}=[h_v^{(1)},\ldots,h_v^{(M)}] \in \mathbb{R}^{M \times d_0}$, hierarchical weights $\omega_m$ combine blur scales:
\begin{equation}
\begin{split}
\tilde{h}_v^{\mathrm{clip}}
&= \sum_{m=1}^{M} \omega_m \odot h_v^{(m)},
\qquad \omega_m = \mathrm{softmax}(\eta_m),
\end{split}
\label{eq:clip_fusion}
\end{equation}
where $\eta_m \in \mathbb{R}^{d_0}$ are hierarchical logits, $\odot$ denotes element-wise multiplication with broadcasting, and $\mathrm{softmax}$ is taken over blur index $m$ for each feature dimension.
For DINO, let $H_{\mathrm{dino}}=[h_v^{(0)},\ldots,h_v^{(L)}]$; the same strategy is applied to layers $\ell=0,\ldots,L$:
\begin{equation}
\begin{split}
\tilde{h}_v^{\mathrm{dino}}
&= \sum_{\ell=0}^{L} \phi_\ell \odot h_v^{(\ell)},
\qquad \phi_\ell = \mathrm{softmax}(\xi_\ell),
\end{split}
\label{eq:dino_layer_fusion}
\end{equation}
with $L{=}14$, where $\xi_\ell \in \mathbb{R}^{d_0}$ are hierarchical logits.
A lightweight MLP then maps the weighted features to teacher embeddings:
\begin{equation}
\begin{aligned}
z_v^{\mathrm{clip}} &= f_H^{\mathrm{clip}}\!\left(H_{\mathrm{clip}}\right)
= \mathrm{MLP}^{\mathrm{clip}}\!\left(\tilde{h}_v^{\mathrm{clip}}\right), \\
z_v^{\mathrm{dino}} &= f_H^{\mathrm{dino}}\!\left(H_{\mathrm{dino}}\right)
= \mathrm{MLP}^{\mathrm{dino}}\!\left(\tilde{h}_v^{\mathrm{dino}}\right).
\end{aligned}
\label{eq:hierarchical_weighting}
\end{equation}

\subsection{Brain Encoder (AMSPBE)}
\label{sec:amspbe_encoder}
\begin{figure}[!t]
  \centering
  \includegraphics[width=\columnwidth]{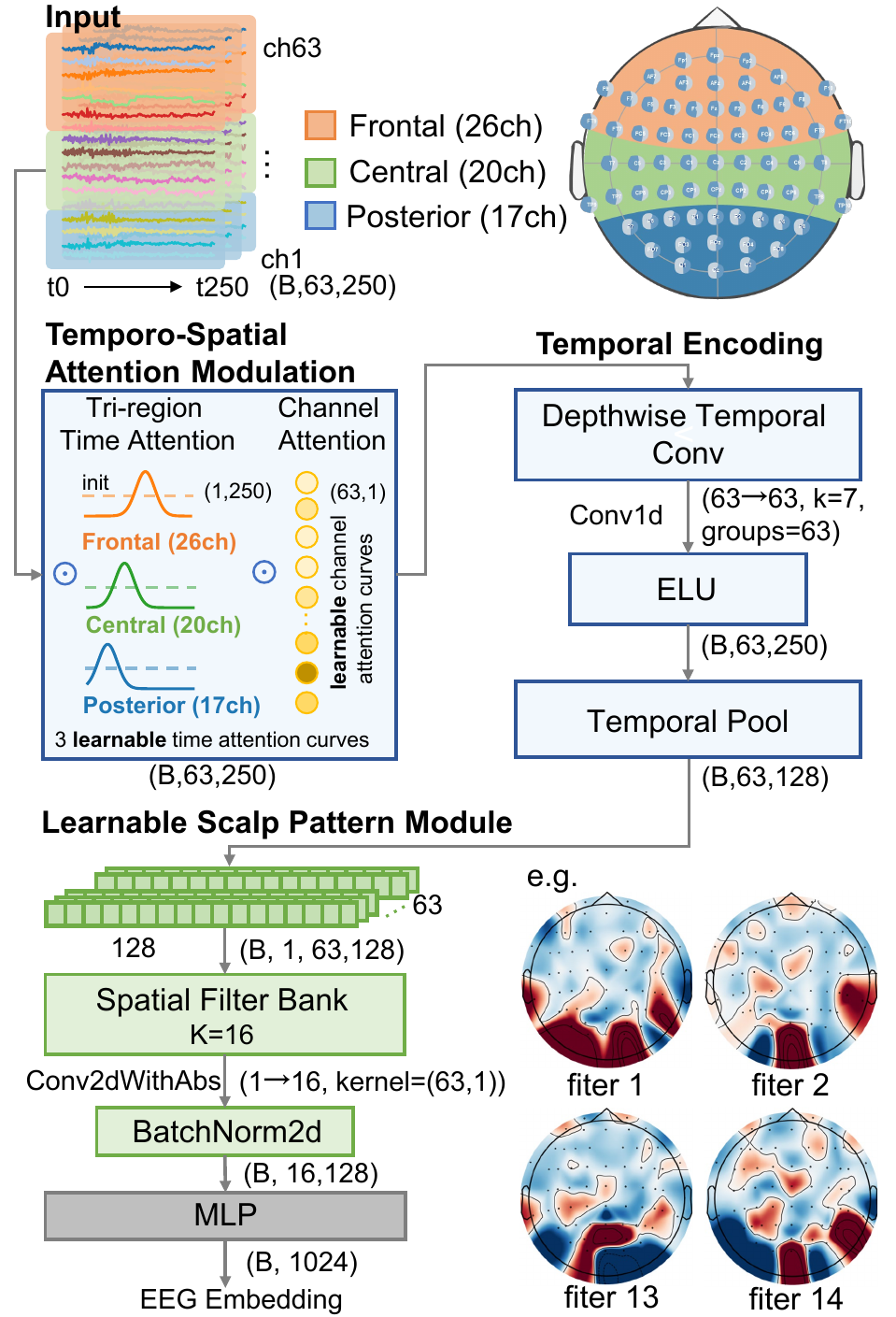}
  \caption{Architecture of AMSPBE ($f_B$). EEG undergoes temporo-spatial attention modulation (TSAM; tri-region temporal attention and per-channel attention), temporal encoding ($k{=}7$, $t_{\mathrm{pool}}{=}128$), and a $K{=}16$ positivity-constrained scalp-pattern filter bank with an MLP projection, yielding brain embedding $z_b$. Right: representative learned scalp-pattern topographies.}
  \label{fig:eeg_encoder}
\end{figure}

As shown in Fig.~\ref{fig:eeg_encoder}, \textbf{AMSPBE} (Attentive Modulated Scalp-Pattern Bank Encoder) is the brain encoder $f_B$, mapping RSVP-evoked EEG to a $1024$-D brain embedding $z_b$ for alignment with the dual-teacher framework.

Under rapid serial visual presentation (RSVP), stimulus images are serially presented to subjects at a rapid rate, so temporally adjacent trials induce \emph{temporal aliasing}~\cite{gifford2022}, leading to significant cross-stimulus interference that degrades single-trial EEG.
Furthermore, brain responses to visual stimuli reflect hierarchical visual processing across scalp regions, and different regions and time points contribute unequally to visual decoding.
Accordingly, we introduce temporo-spatial attention modulation (TSAM) through a coarse tri-region split of the electrode montage into posterior, central, and frontal channel sets: each region shares one learnable temporal attention curve, complemented by per-channel attention (Fig.~\ref{fig:eeg_encoder}).
All attention weights are initialized to one and learned end-to-end to upweight decoding-relevant channels and time points while suppressing aliasing-induced noise.

After attention modulation and lightweight temporal encoding, visual decoding may rely on different collaborative channel combinations across the electrode montage rather than on any single electrode.
Following~\cite{liu2026visualeeg}, we adopt a learnable scalp-pattern filter bank ($K{=}16$) with positivity-constrained spatial convolutions, which learns multiple distinct combination patterns over the electrode layout, as illustrated by several representative scalp-pattern topographies in Fig.~\ref{fig:eeg_encoder}.
These two components---temporo-spatial attention modulation with temporal encoding, and the scalp-pattern filter bank---jointly constitute AMSPBE.
Learned attention curves and filter-bank topographies are analyzed in Sec.~\ref{sec:encoder_analysis}.

\section{Experiments and Analysis}

\subsection{Experimental Setup}
\label{sec:setup_main}

\subsubsection{Datasets and Preprocessing}
We conduct experiments on \textbf{THINGS-EEG}~\cite{gifford2022} and \textbf{THINGS-MEG}~\cite{hebart2023}; dataset scales and acquisition parameters are summarized in Table~\ref{tab:datasets}.
THINGS-EEG provides the primary benchmark for all main results and analyses, while THINGS-MEG is used for cross-modality validation.

The EEG recordings come from ten participants who viewed images under a time-efficient RSVP paradigm.
The training split comprises $1654$ concepts $\times$ $10$ images $\times$ $4$ repetitions, and the test split comprises $200$ concepts $\times$ $1$ image $\times$ $80$ repetitions.
The stimulus onset asynchrony (SOA) is $200$\,ms, consisting of a $100$\,ms image presentation followed by a $100$\,ms blank interval.
For preprocessing, we epoch trials from $0$ to $1000$\,ms after stimulus onset, downsample to 250\,Hz ($T{=}250$), apply baseline correction, and average repetitions within each image to improve the signal-to-noise ratio (SNR).
Following standard THINGS-EEG practice, multivariate noise normalization (MVNN) is applied at the subject level.

The MEG benchmark includes four participants and $271$ channels, with a longer stimulus duration ($500$\,ms) and SOA of $1500{\pm}200$\,ms.
MEG trials are epoched over the same post-stimulus window, band-pass filtered, downsampled to 200\,Hz ($T{=}201$), and MVNN-normalized; configuration and preprocessing follow a similar protocol~\cite{song2025recognizing,wu2025ubp}.
On MEG, the same $f_B$ applies; tri-region channel indices are defined on the CTF-275 electrode montage.

\begin{table}[htbp]
\caption{Configuration details of the THINGS-EEG and THINGS-MEG datasets (Sub: subjects, Chan: channels).}
\label{tab:datasets}
\centering
\small
\setlength{\tabcolsep}{1.1mm}
\begin{tabular}{lccccc}
\toprule
Type & Sub & Chan & Train$^*$ & Test$^*$ & SOA(ms) \\
\midrule
EEG  & 10 & 63  & 1654 $\mid$ 10 $\mid$ 4 & 200 $\mid$ 1 $\mid$ 80 & 200 (100) \\
MEG  & 4  & 271 & 1854 $\mid$ 12 $\mid$ 1 & 200 $\mid$ 1 $\mid$ 12 & 1500$\pm$200 (500) \\
\bottomrule
\end{tabular}
\vspace{2pt}
\raggedright
\noindent\footnotesize $^*$ concepts (classes) $\mid$ conditions (images) $\mid$ repetitions (times).
\end{table}

\subsubsection{Vision Teachers}
Default teachers follow Sec.~\ref{sec:dtha}: frozen CLIP RN50 multi-blur and DINOv2 ViT-L/14 (L0--L14).
Sec.~\ref{sec:teacher_selection} further compares PE-Core and sharp-only CLIP in the image--text paradigm, and DINOv2/v3 teachers across architectures (ViT vs.\ ConvNeXt), scales, and layer spans in the image-only paradigm.

\subsubsection{Brain Encoders}
The default brain encoder is \textbf{AMSPBE} (Sec.~\ref{sec:amspbe_encoder}; Fig.~\ref{fig:eeg_encoder}).
To validate the effectiveness of AMSPBE, we additionally replace it with TSConv~\cite{song2024decoding}, SpatialConv~\cite{liu2026visualeeg}, EEGNet~\cite{lawhern2018eegnet}, ShallowNet/DeepConvNet~\cite{schirrmeister2017deep}, EEGProject~\cite{wu2025ubp}, and STAE~\cite{wu2026ats} under the same dual-teacher training recipe (Sec.~\ref{sec:encoder_analysis}).

\subsubsection{Baselines}
We compare DTHA against recent EEG/MEG visual retrieval methods, including BraVL~\cite{du2023decoding}, NICE~\cite{song2024decoding}, MB2C~\cite{wei2024mb2c}, ATM-S~\cite{li2024visual}, CognitionCapturer~\cite{zhang2025cognitioncapturer}, UBP~\cite{wu2025ubp}, ATS~\cite{wu2026ats}, NeuroCLIP~\cite{wang2025neuroclip}, HVF~\cite{zheng2026hvf}, and LVBP~\cite{liu2026visualeeg}.
Their positioning is discussed in Sec.~\ref{sec:rw_eeg_decoding}.

\subsubsection{Implementation Details}
Models are trained for 250 epochs with batch size 1024 on 8 NVIDIA RTX 3090 GPUs using AdamW ($\mathrm{lr}{=}10^{-3}$, weight decay $10^{-4}$) and a 10-epoch warmup cosine schedule.
Loss weights, temperature, and normalization follow Sec.~\ref{sec:dtha}.
Following standard THINGS-EEG practice~\cite{song2024decoding,liu2026visualeeg}, the official 200-concept test split is reserved for final evaluation only; for model selection and encoder hyperparameter tuning, we randomly hold out 5\% of each subject's training set as a validation split, as in LVBP~\cite{liu2026visualeeg}.
Main results in Tables~\ref{tab:things_eeg_results} and~\ref{tab:things_meg_results} are averaged over seeds $\{2025, 42, 1234\}$; ablations use seed 2025 unless noted otherwise.

\subsection{Main Retrieval Results}
\label{sec:main_results}

\begin{table*}[!t]
    \caption{Top-1 and Top-5 accuracy (\%) for 200-way zero-shot retrieval on THINGS-EEG (10-subject mean; three-seed average).}
    \label{tab:things_eeg_results}
    \centering
    \footnotesize
    \setlength{\tabcolsep}{2.0pt}
    \adjustbox{max width=\textwidth}{%
    \begin{tabular}{@{}lcccccccccccccccccccccc@{}}
    \toprule
    \multirow{2}{*}{Method} & \multicolumn{2}{c}{Subject 1} & \multicolumn{2}{c}{Subject 2} & \multicolumn{2}{c}{Subject 3} & \multicolumn{2}{c}{Subject 4} & \multicolumn{2}{c}{Subject 5} & \multicolumn{2}{c}{Subject 6} & \multicolumn{2}{c}{Subject 7} & \multicolumn{2}{c}{Subject 8} & \multicolumn{2}{c}{Subject 9} & \multicolumn{2}{c}{Subject 10} & \multicolumn{2}{c}{Avg} \\
    \cmidrule(lr){2-3} \cmidrule(lr){4-5} \cmidrule(lr){6-7} \cmidrule(lr){8-9} \cmidrule(lr){10-11} \cmidrule(lr){12-13} \cmidrule(lr){14-15} \cmidrule(lr){16-17} \cmidrule(lr){18-19} \cmidrule(lr){20-21} \cmidrule(lr){22-23}
    & top-1 & top-5 & top-1 & top-5 & top-1 & top-5 & top-1 & top-5 & top-1 & top-5 & top-1 & top-5 & top-1 & top-5 & top-1 & top-5 & top-1 & top-5 & top-1 & top-5 & top-1 & top-5 \\
    \midrule
    \multicolumn{23}{c}{\textbf{Intra-subject}: train and test on one subject} \\
    \midrule
    \tblmethod{BraVL}{du2023decoding}{TPAMI'23} & 6.1 & 17.9 & 4.9 & 14.9 & 5.6 & 17.4 & 5.0 & 15.1 & 4.0 & 13.4 & 6.0 & 18.2 & 6.5 & 20.4 & 8.8 & 23.7 & 4.3 & 14.0 & 7.0 & 19.7 & 5.8 & 17.5 \\
    \tblmethod{NICE}{song2024decoding}{ICLR'24} & 13.2 & 39.5 & 13.5 & 40.3 & 14.5 & 42.7 & 20.6 & 52.7 & 10.1 & 31.5 & 16.5 & 44.0 & 17.0 & 42.1 & 22.9 & 56.1 & 15.4 & 41.6 & 17.4 & 45.8 & 16.1 & 43.6 \\
    \tblmethod{MB2C}{wei2024mb2c}{ACM MM'24} & 23.6 & 56.3 & 22.6 & 50.5 & 26.3 & 60.1 & 34.8 & 67.0 & 21.3 & 53.0 & 31.0 & 62.3 & 25.0 & 54.8 & 39.0 & 69.3 & 27.5 & 59.3 & 33.1 & 70.8 & 28.4 & 60.3 \\
    \tblmethod{ATM-S}{li2024visual}{NeurIPS'24} & 25.6 & 60.4 & 22.0 & 54.5 & 24.0 & 62.4 & 31.4 & 60.9 & 12.9 & 43.0 & 21.4 & 51.1 & 30.5 & 61.5 & 38.8 & 72.0 & 34.4 & 51.5 & 29.1 & 63.5 & 28.5 & 60.4 \\
    \tblmethod{CogCap}{zhang2025cognitioncapturer}{AAAI'25} & 31.4 & 79.6 & 31.4 & 77.8 & 38.1 & 85.6 & 40.3 & 85.8 & 24.4 & 66.3 & 34.8 & 78.7 & 34.6 & 80.9 & 48.1 & 88.6 & 37.4 & 79.3 & 35.5 & 79.2 & 35.6 & 80.2 \\
    \tblmethod{UBP}{wu2025ubp}{CVPR'25} & 41.2 & 70.5 & 51.2 & 80.9 & 51.2 & 82.0 & 51.1 & 76.9 & 42.2 & 72.8 & 57.5 & 83.5 & 49.0 & 79.9 & 58.6 & 85.8 & 45.1 & 76.2 & 61.5 & 88.2 & 50.9 & 79.7 \\
    \tblmethod{ATS}{wu2026ats}{AAAI'26} & 53.0 & 79.0 & 62.0 & 87.5 & 61.5 & 89.0 & 57.0 & 86.5 & 55.0 & 84.0 & 68.0 & 90.5 & 53.0 & 84.0 & 66.5 & 91.0 & 58.5 & 86.0 & 67.5 & 89.0 & 60.2 & 86.7 \\
    \tblmethod{NeuroCLIP}{wang2025neuroclip}{arXiv'26} & 60.3 & 84.2 & 64.5 & 94.1 & 63.7 & 93.2 & 62.8 & 91.3 & 55.0 & 83.7 & 68.6 & 93.6 & 70.6 & 93.3 & 69.6 & 90.3 & 48.2 & 86.5 & 69.1 & 92.1 & 63.2 & 90.3 \\
    \tblmethod{HVF}{zheng2026hvf}{ICLR'26} & 64.3 & 88.8 & 76.3 & 95.3 & 74.0 & 95.0 & 67.0 & 91.8 & 68.0 & 91.5 & 81.5 & 96.3 & 76.8 & 96.8 & 84.8 & 98.5 & 76.8 & 95.8 & 87.3 & 99.3 & 75.7 & 94.6 \\
    \tblmethod{LVBP}{liu2026visualeeg}{AAAI'26} & 81.9 & 96.8 & 81.7 & 96.8 & 78.3 & 96.0 & 76.9 & 97.2 & 71.4 & 93.7 & 84.3 & 98.7 & 78.2 & 96.7 & 84.5 & 98.1 & 78.7 & 96.8 & 84.1 & 98.1 & 80.0 & 96.9 \\
    \midrule
    \rowcolor{gray!15}
    \textbf{DTHA (ours)} & \textbf{94.7} & \textbf{99.0} & \textbf{90.7} & \textbf{99.8} & \textbf{91.3} & \textbf{100.0} & \textbf{90.3} & \textbf{99.7} & \textbf{87.8} & \textbf{99.0} & \textbf{94.8} & \textbf{99.5} & \textbf{91.5} & \textbf{98.7} & \textbf{93.2} & \textbf{99.5} & \textbf{94.2} & \textbf{99.5} & \textbf{95.2} & \textbf{100.0} & \textbf{92.4} & \textbf{99.5} \\
    \midrule
    \multicolumn{23}{c}{\textbf{Inter-subject}: leave one subject out for test} \\
    \midrule
    \tblmethod{BraVL}{du2023decoding}{TPAMI'23} & 2.3 & 8.0 & 1.5 & 6.3 & 1.4 & 5.9 & 1.7 & 6.7 & 1.5 & 5.6 & 1.8 & 7.2 & 2.1 & 8.1 & 2.2 & 7.6 & 1.6 & 6.4 & 2.3 & 8.5 & 1.8 & 7.0 \\
    \tblmethod{NICE}{song2024decoding}{ICLR'24} & 7.6 & 22.8 & 5.9 & 20.5 & 6.0 & 22.3 & 6.3 & 20.7 & 4.4 & 18.3 & 5.6 & 22.2 & 5.6 & 19.7 & 6.3 & 22.0 & 5.7 & 17.6 & 8.4 & 28.3 & 6.2 & 21.4 \\
    \tblmethod{ATM-S}{li2024visual}{NeurIPS'24} & 10.5 & 26.8 & 7.1 & 24.8 & 11.9 & 33.8 & 14.7 & 39.4 & 7.0 & 23.9 & 11.1 & 35.8 & 16.1 & 43.5 & 15.0 & 40.3 & 4.9 & 22.7 & 20.5 & 46.5 & 11.8 & 33.7 \\
    \tblmethod{UBP}{wu2025ubp}{CVPR'25} & 11.5 & 29.7 & 15.5 & 40.0 & 9.8 & 27.0 & 13.0 & 32.3 & 8.8 & 33.8 & 11.7 & 31.0 & 10.2 & 23.8 & 12.2 & 32.2 & 15.5 & 40.5 & 16.0 & 43.5 & 12.4 & 33.4 \\
    \tblmethod{ATS}{wu2026ats}{AAAI'26} & 17.0 & 37.5 & 21.5 & 49.5 & 7.5 & 20.0 & 13.5 & 41.5 & 7.0 & 27.0 & 14.5 & 37.0 & 10.0 & 33.5 & 13.0 & 30.5 & 9.5 & 28.5 & 26.0 & 52.5 & 14.0 & 35.8 \\
    \tblmethod{NeuroCLIP}{wang2025neuroclip}{arXiv'26} & 17.2 & 47.3 & 31.8 & 60.2 & 10.9 & 24.3 & 19.1 & 48.2 & 14.3 & 32.6 & 11.2 & 32.1 & 15.8 & 41.3 & 10.2 & 29.2 & 9.9 & 27.4 & 29.8 & 60.0 & 17.0 & 40.3 \\
    \tblmethod{LVBP}{liu2026visualeeg}{AAAI'26} & 27.3 & 54.9 & 32.0 & 63.0 & 11.1 & 34.5 & 18.6 & 46.0 & 16.9 & 40.7 & 16.1 & 43.2 & 16.0 & 45.4 & 18.3 & 49.0 & 13.6 & 41.2 & 30.1 & 62.1 & 20.0 & 48.0 \\
    \midrule
    \rowcolor{gray!15}
    \textbf{DTHA (ours)} & \textbf{31.3} & \textbf{65.3} & \textbf{39.0} & \textbf{69.5} & \textbf{14.3} & \textbf{38.7} & \textbf{23.0} & \textbf{55.3} & \textbf{21.2} & \textbf{42.7} & \textbf{23.8} & \textbf{57.8} & \textbf{25.8} & \textbf{54.7} & \textbf{20.8} & \textbf{48.7} & \textbf{17.5} & \textbf{43.8} & \textbf{39.2} & \textbf{73.2} & \textbf{25.6} & \textbf{55.0} \\
    \bottomrule
    \end{tabular}}
  \end{table*}

Table~\ref{tab:things_eeg_results} reports intra- and inter-subject 200-way retrieval on THINGS-EEG.
DTHA achieves \textbf{92.4\%} Top-1 and \textbf{99.5\%} Top-5 accuracy, surpassing the previous state of the art by \textbf{12.4} and \textbf{2.6} percentage points, respectively.
Across subjects, Top-1 is \textbf{92.4\% $\pm$ 2.4\%} with a bootstrap 95\% CI of \textbf{[90.9\%, 93.8\%]} ($n{=}10$).
A paired Wilcoxon signed-rank test on per-subject Top-1 scores confirms that DTHA significantly outperforms LVBP ($p < 0.001$), with a large paired effect size (Cohen's $d_z = 3.76$).
\begin{table}[!t]
    \caption{Top-1 and Top-5 accuracy (\%) for 200-way zero-shot retrieval on THINGS-MEG (four-subject mean; three-seed average).}
    \label{tab:things_meg_results}
    \centering
    \small
    \setlength{\tabcolsep}{1.4pt}
    \begin{tabular}{lcccccccccc}
      \toprule
      \multirow{2}{*}{Method}& \multicolumn{2}{c}{Subject 1} & \multicolumn{2}{c}{Subject 2} & \multicolumn{2}{c}{Subject 3} & \multicolumn{2}{c}{Subject 4}& \multicolumn{2}{c}{Avg} \\
      \cmidrule(lr){2-3} \cmidrule(lr){4-5} \cmidrule(lr){6-7} \cmidrule(lr){8-9} \cmidrule(lr){10-11}
      & top-1 & top-5 & top-1 & top-5 & top-1 & top-5 & top-1 & top-5 & top-1 & top-5 \\
      \midrule
      \multicolumn{11}{c}{\textbf{Intra-subject}: train and test on one subject} \\
      \midrule
      NICE & 9.6 & 27.8 & 18.5 & 47.8 & 14.2 & 41.6 & 9.0 & 26.6 & 12.8 & 36.0 \\
      UBP & 15.0 & 38.0 & 46.0 & 80.5 & 27.3 & 59.0 & 18.5 & 43.5 & 26.7 & 55.2 \\
      ATS & 18.0 & 43.0 & 56.0 & 89.0 & 36.0 & 67.0 & 19.5 & 50.0 & 32.4 & 62.3 \\
      HVF & 14.0 & 31.8 & 63.8 & 91.8 & 41.0 & 78.3 & 17.0 & 41.0 & 33.9 & 60.7 \\
      LVBP & 27.0 & 59.8 & 75.7 & 96.2 & 50.3 & 82.8 & 34.0 & 66.7 & 46.8 & 76.4 \\
            \midrule
      \rowcolor{gray!15}
      \textbf{DTHA} & \textbf{39.8} & \textbf{69.7} & \textbf{90.2} & \textbf{99.7} & \textbf{75.0} & \textbf{96.0} & \textbf{52.5} & \textbf{82.0} & \textbf{64.4} & \textbf{86.8} \\
      \midrule
      \multicolumn{11}{c}{\textbf{Inter-subject}: leave one subject out for test} \\
      \midrule
      UBP & 2.0 & 5.7 & 1.5 & 17.2 & 2.7 & 10.5 & 2.5 & 8.0 & 2.2 & 10.4 \\
      ATS & 2.0 & 6.0 & 6.0 & 14.0 & 4.5 & 15.0 & 1.0 & 9.5 & 3.4 & 11.2 \\
      LVBP & 1.0 & 6.5 & 6.5 & 19.5 & 7.3 & 20.8 & 3.3 & 13.7 & 4.5 & 15.1 \\
      \midrule
      \rowcolor{gray!15}
      \textbf{DTHA} & \textbf{1.5} & \textbf{9.3} & \textbf{8.3} & \textbf{25.2} & \textbf{8.8} & \textbf{27.7} & \textbf{3.8} & \textbf{13.7} & \textbf{5.6} & \textbf{19.0} \\
      \bottomrule
    \end{tabular}
  \end{table}
 Table~\ref{tab:things_meg_results} reports intra- and inter-subject 200-way retrieval on THINGS-MEG.
DTHA achieves \textbf{64.4\%} Top-1 and \textbf{86.8\%} Top-5 accuracy, improving over LVBP by \textbf{17.6} and \textbf{10.4} percentage points (mean Top-1: \textbf{64.4\% $\pm$ 22.5\%}; 95\% CI: \textbf{[46.2\%, 82.6\%]}; $n{=}4$ subjects; $d_z = 1.73$).

Substantial improvements on both benchmarks support DTHA's effectiveness and generalization across modalities. The following analyses examine the motivation and design rationale behind complementarity-driven dual-teacher alignment.

\subsection{Cross-Paradigm Teacher Representational Analysis}
\label{sec:teacher_repr_diff}

To assess our mechanism--paradigm correspondence view, we first examine whether cross-paradigm CLIP and DINO teachers provide differentiated supervision for EEG-based visual decoding, and how each should be configured to emphasize the visual cues its training paradigm best supports.

\subsubsection{Semantic vs.\ Appearance-Related Structure}
We define six top-down categories by aggregating THINGS manual metadata (food, animal, container, clothing, tool, and vehicle) and perform coarse-category representational similarity analysis (RSA)~\cite{kriegeskorte2008} on THINGS-EEG training images (494 concepts, one image per concept; Fig.~\ref{fig:clip_dino_repr}(A)).
CLIP-RN50 exhibits a clear block-diagonal pattern aligned with these categories, indicating that its representation space is organized by category-level concepts.
DINOv2 ViT-L/14, by contrast, shows much weaker category clustering and no comparable within-category structure.
This contrast supports differentiated cross-paradigm supervision roles---CLIP as a \emph{semantic} teacher and DINO as an \emph{appearance-related} teacher---rather than two redundant encoders with overlapping representational biases.

\subsubsection{Multi-Blur CLIP Supervision}
\begin{figure}[!t]
  \centering
  \includegraphics[width=\columnwidth]{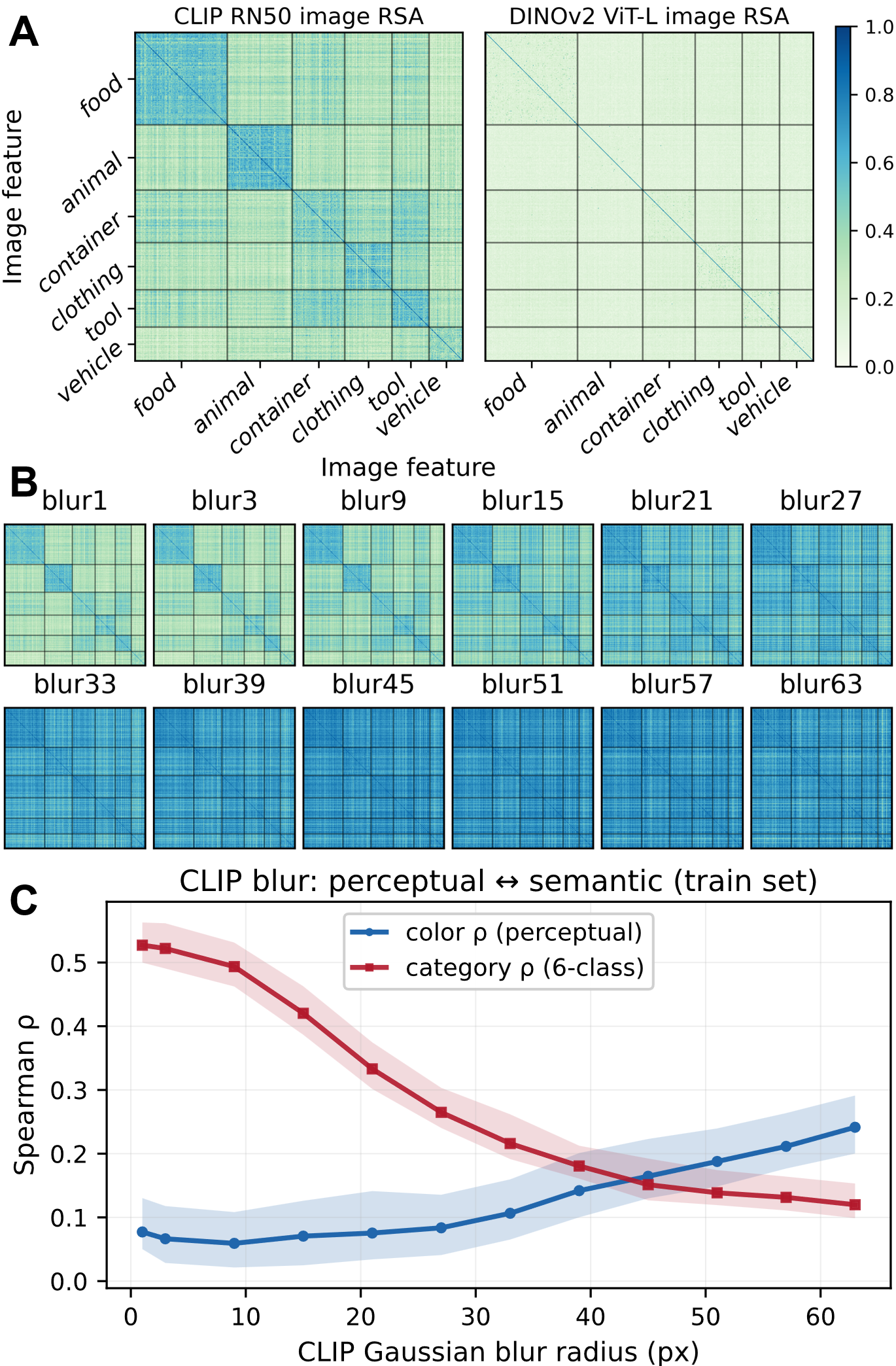}
  \caption{CLIP vs.\ DINO representational differences on THINGS-EEG \textbf{training} images.
  (A) Category-semantic RSA for CLIP-RN50 and DINOv2 ViT-L/14: CLIP shows block-diagonal clustering; DINO does not.
  (B) CLIP category-semantic RSA at each blur level ($\sigma{=}1$--$63$): block-diagonal clustering weakens as $\sigma$ increases.
  (C) Spearman $\rho$ of CLIP RDMs with category-semantic and chroma reference RDMs across blur levels.}
  \label{fig:clip_dino_repr}
\end{figure}

Following LVBP~\cite{liu2026visualeeg}, we adopt its multi-level blur on image inputs but interpret its role differently.
Fig.~\ref{fig:clip_dino_repr}(B) shows that block-diagonal category-semantic clustering weakens progressively as $\sigma$ increases.
Fig.~\ref{fig:clip_dino_repr}(C) tracks this shift numerically by asking, at each blur level, whether CLIP groups images more by category semantics or by color similarity.
For each $\sigma$, we summarize pairwise image dissimilarities as representational dissimilarity matrices (RDMs; pairwise stimulus dissimilarities) in CLIP space and in two reference spaces---category-semantic proxies and Lab-chroma---and compute Spearman rank correlation $\rho$ between them over all image pairs; $\rho$ measures rank-order consistency between the CLIP and reference distance structures---higher $\rho$ means pairs that are more dissimilar under the reference also tend to be more dissimilar in CLIP space.
Fig.~\ref{fig:clip_dino_repr}(C) shows that stronger blur weakens category-semantic consistency while strengthening chroma consistency, yet category-semantic $\rho$ remains substantially higher.
Accordingly, multi-blur CLIP supplements this semantic branch with blur-induced appearance-related information.

\subsubsection{Shallow DINO Supervision}
\begin{figure}[!t]
  \centering
  \includegraphics[width=\columnwidth]{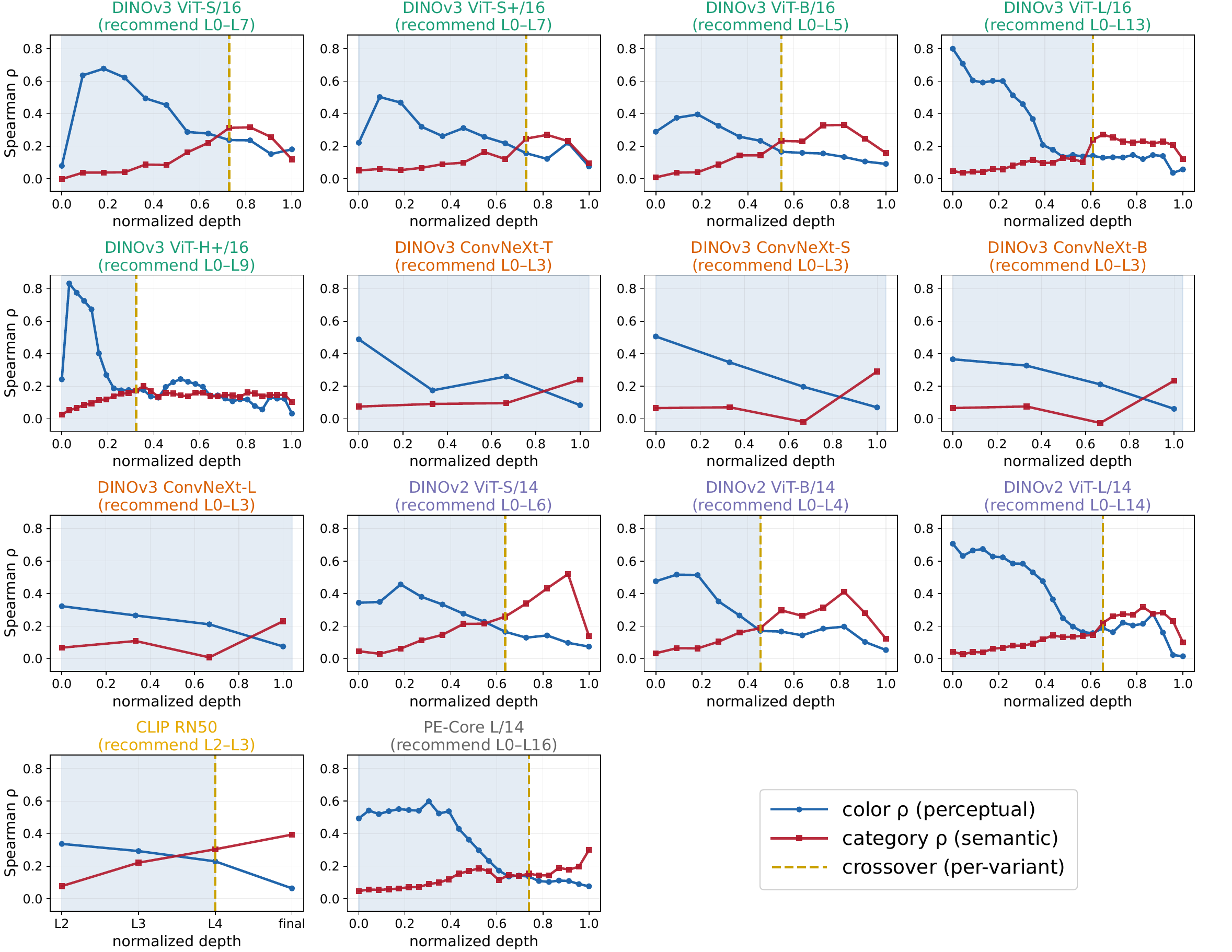}
  \caption{Layer-wise Spearman $\rho$ of teacher RDMs with category-semantic and chroma reference RDMs (CLIP, PE-Core, DINO; THINGS-EEG \textbf{test}, $n{=}200$).
  Gold dashed lines mark semantic--chroma $\rho$ crossovers.}
  \label{fig:grid_color_category}
\end{figure}

\begin{figure}[!t]
  \centering
  \includegraphics[width=\columnwidth]{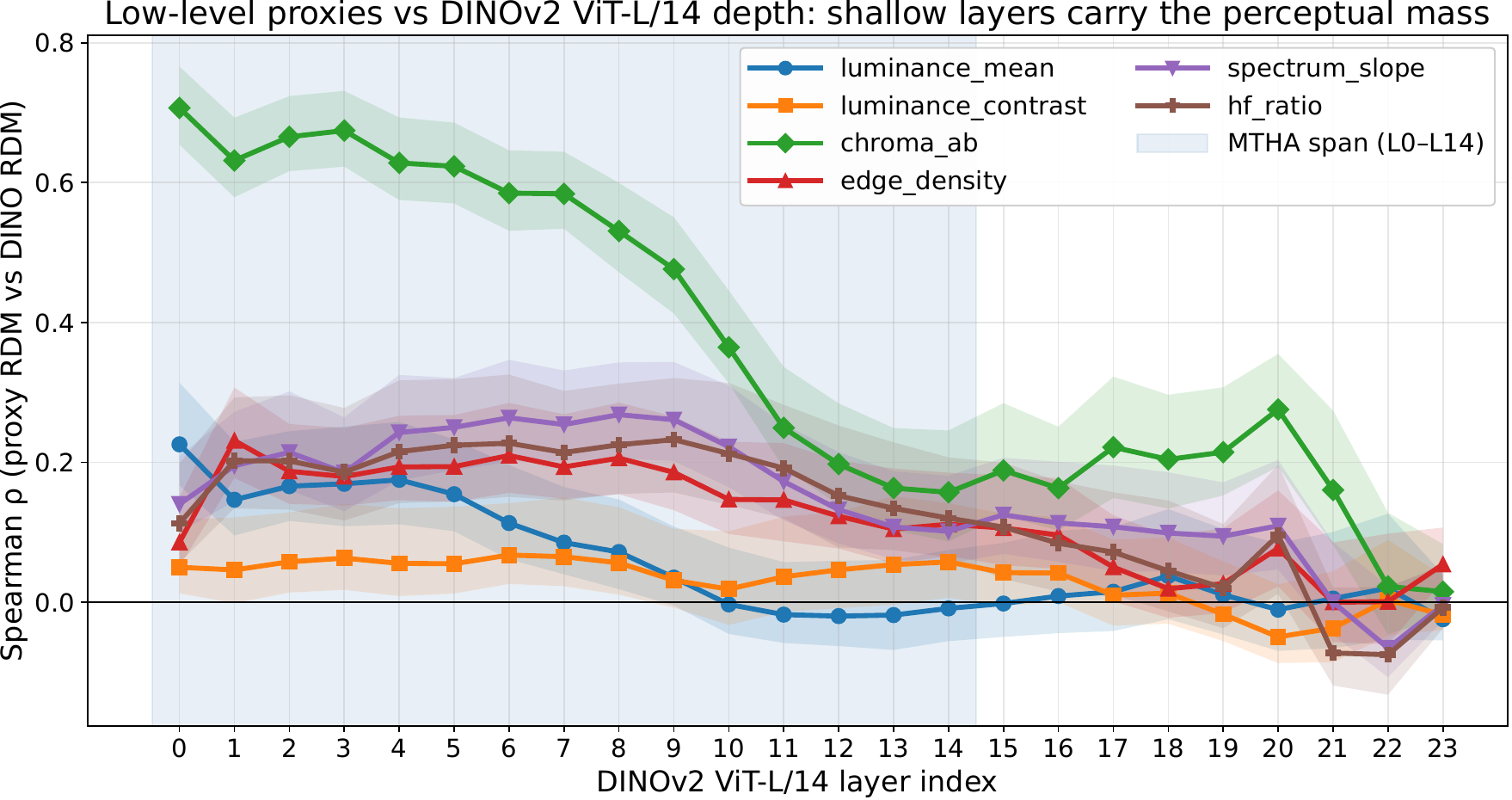}
  \caption{Layer-wise Spearman $\rho$ of DINOv2 ViT-L/14 RDMs with six low-level reference RDMs---chroma, luminance, edge density, and spectral statistics---(THINGS-EEG \textbf{test}, $n{=}200$). Shallow layers show generally higher $\rho$.}
  \label{fig:lowlevel_proxies}
\end{figure}

Because multi-blur CLIP supplies only partial appearance-related cues on the semantic branch, we turn to DINO for richer structure-related supervision and examine which depth range best preserves appearance-related structure.
On THINGS-EEG test images ($n{=}200$ concepts), Fig.~\ref{fig:grid_color_category} compares Spearman $\rho$ with chroma and category-semantic proxies across CLIP, PE-Core, and multiple DINO variants by depth.
Chroma consistency peaks in shallow layers and weakens with depth, whereas category-semantic consistency grows in deeper layers; on most ViT teachers the semantic and chroma $\rho$ curves cross (gold dashed lines).
Fig.~\ref{fig:lowlevel_proxies} adds six low-level visual proxies (chroma, luminance, edge density, and spectral statistics) to corroborate this depth trend: all show generally higher $\rho$ in shallow layers and weaken with depth, with layer~14 near the crossover.
Accordingly, we adopt shallow DINOv2 layers (L0--L14) as the appearance-related teacher.

\subsection{Vision Teacher Selection}
\label{sec:teacher_selection}

From representational analyses in Sec.~\ref{sec:teacher_repr_diff}, we specify multi-blur CLIP as the semantic teacher and shallow DINO layers as the appearance-related teacher; we next validate each teacher through single-teacher retrieval sweeps.
With AMSPBE and all other training settings held constant, each sweep uses only the CLIP or DINO branch SCE loss (Eq.~\eqref{eq:branch_loss}), without the fused-retrieval objective (Fig.~\ref{fig:teacher_single_sweep}).

\subsubsection{CLIP Teachers and Multi-Blur}
As the top of Fig.~\ref{fig:teacher_single_sweep} shows, CLIP-style teachers consistently outperform PE-Core on the semantic branch.
Multi-blur supervision exceeds sharp-only inputs ($\sigma{=}1$) for both, consistent with Sec.~\ref{sec:teacher_repr_diff}: blur supplements appearance-related cues beyond sharp image--text features alone; RN50 multi-blur reaches the strongest branch Top-1 ($79.7\%$).
We do not sweep CLIP architectures extensively; our prior ATS work~\cite{wu2026ats} shows that overly strong vision teachers widen the brain--vision asymmetry, and RN50 offers the best trade-off among the teachers tested.

\subsubsection{DINO Teachers and Depth Span}
\begin{figure}[!t]
  \centering
  \includegraphics[width=\columnwidth]{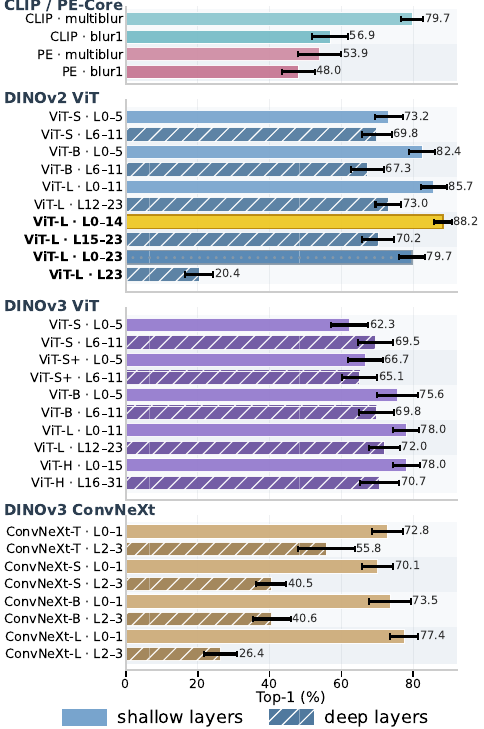}
  \caption{Single-teacher retrieval sweep (branch Top-1, \%; 10-subject mean, seed 2025).
  Each condition uses branch SCE only (Eq.~\eqref{eq:branch_loss}), without fused-retrieval.
  Top: CLIP/PE-Core blur sweeps; bottom: DINOv2/v3 layer spans on ViT and ConvNeXt. Solid/hatched bars mark shallow/deep mid-depth splits.}
  \label{fig:teacher_single_sweep}
\end{figure}

On the appearance-related branch, we compare DINOv2 and DINOv3 across architectures, scales, and layer spans.
DINOv2 ViT-L/14 yields the strongest single-teacher retrieval.
Although DINOv3 generally leads on vision benchmarks, matched-scale DINOv3 teachers underperform DINOv2 on EEG retrieval in this sweep.
DINOv3's Gram anchoring~\cite{simeoni2025dinov3} stabilizes dense patch features for geometric vision tasks, but the resulting patch-level features carry finer local visual structure than noisy scalp EEG can reliably encode; consistent with the brain--vision asymmetry in our prior ATS work~\cite{wu2026ats}, this mismatch widens the modality gap and hurts retrieval.
For layer-span comparisons in Fig.~\ref{fig:teacher_single_sweep}, we adopt a uniform mid-depth split (first vs.\ second half of blocks; solid vs.\ hatched bars); shallow spans generally outperform deep spans, corroborating Sec.~\ref{sec:teacher_repr_diff}.
Motivated by the crossover near layer~14 in Fig.~\ref{fig:lowlevel_proxies}, we further test DINOv2 ViT-L/14 with L0--L14: it exceeds the half-depth shallow span (L0--L11), deep layers L15--L23, and full-depth fusion (L0--L23), while the final-layer CLS reaches only $20.4\%$ (Fig.~\ref{fig:teacher_single_sweep}).
This contrast further underscores why the layer-wise DINO analysis in Sec.~\ref{sec:teacher_repr_diff} is essential for EEG-based visual decoding, rather than conventionally adopting the final-layer CLS.
These sweeps establish our default pairing: CLIP RN50 multi-blur + DINOv2 ViT-L/14 (L0--L14).

\begin{table*}[!t]
  \caption{Representative dual-teacher pairings on THINGS-EEG (10-subject Top-1, \%; seed 2025).
  Single$_A$/Single$_B$/Dual$_A$/Dual$_B$: Top-1 under single-teacher vs full DTHA training.
  $\Delta_A{=}\mathrm{Dual}_A{-}\mathrm{Single}_A$, $\Delta_B{=}\mathrm{Dual}_B{-}\mathrm{Single}_B$, $\Delta_{\mathrm{co}}{=}\Delta_A{+}\Delta_B$, $\Delta_{\mathrm{br}}{=}|\mathrm{Dual}_B{-}\mathrm{Dual}_A|$.
  $\Delta_{\mathrm{gain}}$: fused minus $\max($Single$_A$, Single$_B$, Dual$_A$, Dual$_B)$; \textit{Effective}: \cmark\ ($>{+}0.5$\,pp), \xmark\ ($<-0.5$\,pp), else $\sim$.}
  \label{tab:teacher_fusion}
  \centering
  \scriptsize
  \setlength{\tabcolsep}{2.6pt}
  \adjustbox{width=\textwidth}{%
  \begin{tabular}{@{} c l l l *{11}{c} @{}}
    \toprule
    & & & & \multicolumn{2}{c}{Single-teacher} & \multicolumn{3}{c}{Dual-trained} & \multicolumn{3}{c}{Synergy} & \multicolumn{3}{c}{Fusion} \\
    \cmidrule(lr){5-6}\cmidrule(lr){7-9}\cmidrule(lr){10-12}\cmidrule(lr){13-15}
    \# & Type & \multicolumn{1}{c}{Branch A} & \multicolumn{1}{c}{Branch B} & Single$_A$ & Single$_B$ & Dual$_A$ & Dual$_B$ & $\Delta_{\mathrm{br}}$ & $\Delta_A$ & $\Delta_B$ & $\Delta_{\mathrm{co}}$ & Fused & $\Delta_{\mathrm{gain}}$ & Effective \\
    \midrule
    1 & Cross-modal (main) & \textbf{CLIP RN50 multi-blur} & \textbf{DINOv2 L0-14} & 79.7 & 88.2 & 81.2 & 89.0 & 7.8 & +1.5 & \textbf{+0.8} & +2.3 & \textbf{92.6} & +3.5 & \cmark \\
    \addlinespace[2pt]
    2 & Cross-modal & CLIP RN50 multi-blur & DINOv3 L0-11 & 79.7 & 78.0 & 80.9 & 78.7 & 2.2 & +1.2 & +0.7 & +1.9 & 89.0 & \textbf{+8.0} & \cmark \\
    \addlinespace[2pt]
    3 & Weak branch & CLIP RN50 multi-blur & PE-Core L/14 multi-blur & 79.7 & 53.9 & 79.0 & 56.0 & 23.0 & $-$0.7 & +2.1 & +1.4 & 77.4 & $-$2.3 & \xmark \\
    4 & Weak branch & PE-Core L/14 multi-blur & DINOv2 L0-14 & 53.9 & 88.2 & 56.5 & 86.0 & 29.5 & +2.6 & $-$2.2 & +0.4 & 83.5 & $-$4.8 & \xmark \\
    5 & Weak branch & PE-Core L/14 multi-blur & DINOv3 L0-14 & 53.9 & 78.0 & 56.2 & 78.8 & 22.5 & +2.3 & +0.8 & +3.1 & 79.2 & +0.4 & $\sim$ \\
    \addlinespace[2pt]
    6 & Redundant & DINOv3 L0-14 & DINOv2 L0-14 & 78.0 & 88.2 & 79.0 & 89.0 & 10.0 & +1.0 & +0.8 & +1.8 & 88.0 & $-$1.0 & \xmark \\
    7 & Redundant & DINOv3 L12-23 & DINOv2 L12-23 & 72.0 & 73.0 & 71.3 & 72.2 & 0.9 & $-$0.7 & $-$0.8 & $-$1.5 & 79.6 & +6.5 & \cmark \\
    \addlinespace[2pt]
    8 & Deep ablation & CLIP RN50 multi-blur & DINOv2 L15-23 & 79.7 & 70.2 & 81.2 & 70.5 & 10.6 & +1.5 & +0.3 & +1.8 & 83.0 & +1.8 & \cmark \\
    9 & Deep ablation & CLIP RN50 multi-blur & DINOv3 L12-23 & 79.7 & 72.0 & 80.0 & 72.3 & 7.6 & +0.3 & +0.3 & +0.6 & 83.8 & +3.8 & \cmark \\
    \addlinespace[2pt]
    10 & Shallow + deep & DINOv2 L0-11 & DINOv2 L12-23 & 85.7 & 73.0 & 85.5 & 72.3 & 13.2 & $-$0.2 & $-$0.7 & $-$0.9 & 85.8 & +0.1 & $\sim$ \\
    \bottomrule
  \end{tabular}%
  }
\end{table*}

\subsection{Dual-Teacher Fusion Retrieval}
\label{sec:dual_teacher_valid}

Secs.~\ref{sec:teacher_repr_diff} and~\ref{sec:teacher_selection} establish our default cross-paradigm teachers; we next evaluate full DTHA training (branch SCE plus fused-retrieval).
We ask whether fused retrieval exceeds either teacher alone, which teacher pairings yield fusion gain for EEG-based visual decoding, and how the training objective and fusion architecture should be configured.
Tables~\ref{tab:dual_teacher_overall}--\ref{tab:fusion_training_ablation} report the results for these three aspects, respectively.

\subsubsection{Dual-Teacher vs.\ Single-Teacher Retrieval}
Table~\ref{tab:dual_teacher_overall} compares DTHA fused retrieval with matched single-teacher CLIP and DINO baselines on THINGS-EEG and THINGS-MEG under the default teachers (Sec.~\ref{sec:teacher_selection}).
In Part~I, fused Top-1 reaches 92.4\% (EEG) and 64.4\% (MEG), surpassing CLIP-only by 11.8/12.3~pp and DINO-only by 3.7/7.7~pp.
In Part~II, \emph{branch retrieval} with $S^{\mathrm{clip}}$ or $S^{\mathrm{dino}}$ alone under joint DTHA training already beats the matching Part~I single-teacher run (EEG: $+0.6$/$+0.3$~pp; MEG: $+1.8$/$+1.1$~pp), reflecting co-training synergy under complementary supervision.
The Part~I fused score further exceeds both Part~II branch retrievals on EEG and MEG, confirming fusion gain beyond either the semantic or appearance-related branch alone and indicating that learned similarity-level fusion complementarily combines both supervision axes.

\begin{table}[!t]
  \caption{Overall dual-teacher comparison on THINGS-EEG (10-subject mean) and THINGS-MEG (4-subject mean); three-seed averages.
  Part~I: fused $S_{\mathrm{fused}}$ vs.\ single-teacher CLIP/DINO (branch SCE only); Part~II: $S^{\mathrm{clip}}$/$S^{\mathrm{dino}}$ branch retrieval under full DTHA.
  $\Delta_{\mathrm{fused}}$: row Top-1 $-$ DTHA fused; $\Delta_{\mathrm{single}}$: Part~II $-$ matching Part~I single-teacher row.}
  \label{tab:dual_teacher_overall}
  \centering
  \footnotesize
  \setlength{\tabcolsep}{2.5pt}
  \begin{tabular}{@{}l ccc ccc@{}}
    \toprule
    \addlinespace[2pt]
    \multicolumn{7}{@{}l@{}}{\textit{I.\ Single-teacher training vs.\ DTHA fused retrieval}} \\
    \addlinespace[1pt]
    \multirow{2}{*}{Setting} & \multicolumn{3}{c}{THINGS-EEG} & \multicolumn{3}{c}{THINGS-MEG} \\
    \cmidrule(lr){2-4}\cmidrule(lr){5-7}
    & Top-1 & Top-5 & $\Delta_{\mathrm{fused}}$
    & Top-1 & Top-5 & $\Delta_{\mathrm{fused}}$ \\
    \midrule
    CLIP (single-teacher) & 80.6 & 96.9 & $-11.8$ & 52.1 & 81.3 & $-12.3$ \\
    DINO (single-teacher) & 88.7 & 99.2 & $-3.7$ & 56.7 & 84.0 & $-7.7$ \\
    \textbf{DTHA fused (default)} & \textbf{92.4} & \textbf{99.5} & --
    & \textbf{64.4} & \textbf{86.8} & -- \\
    \midrule
    \addlinespace[2pt]
    \multicolumn{7}{@{}l@{}}{\textit{II.\ Per-branch retrieval: DTHA joint training vs.\ single-teacher training}} \\
    \addlinespace[1pt]
    \multirow{2}{*}{Setting} & \multicolumn{3}{c}{THINGS-EEG} & \multicolumn{3}{c}{THINGS-MEG} \\
    \cmidrule(lr){2-4}\cmidrule(lr){5-7}
    & Top-1 & Top-5 & $\Delta_{\mathrm{single}}$
    & Top-1 & Top-5 & $\Delta_{\mathrm{single}}$ \\
    \midrule
    CLIP branch retrieval & 81.2 & 97.2 & $+0.6$ & 53.9 & 81.4 & $+1.8$ \\
    DINO branch retrieval & 89.0 & 99.1 & $+0.3$ & 57.8 & 83.9 & $+1.1$ \\
    \bottomrule
  \end{tabular}
\end{table}

\subsubsection{Cross-Paradigm Teacher Pairings}
Table~\ref{tab:teacher_fusion} compares representative teacher pairings under full DTHA on THINGS-EEG.
Cross-paradigm CLIP+DINO rows (1--2) yield positive $\Delta_{\mathrm{gain}}$ throughout; row~2 is notable---with weak DINOv3 L0--11 (78.0\% single-teacher Top-1), fused retrieval still reaches 89.0\% ($\Delta_{\mathrm{gain}}{=}{+}8.0$~pp), validating complementary fusion even when one branch is individually weak.
Rows~3--5 show poor fusion performance when PE-Core underperforms on the semantic branch ($\Delta_{\mathrm{gain}}$ from $-4.8$ to $+0.4$~pp; $\Delta_{\mathrm{br}}$ up to 29.5), as a collapsed image--text teacher cannot sustain reliable cross-paradigm supervision.
Same-paradigm stacking also underperforms: row~6 (DINOv3+DINOv2 shallow) yields $\Delta_{\mathrm{gain}}{=}{-}1.0$~pp, and row~10 (DINOv2 L0--11 plus L12--23) reaches only $+0.1$~pp without sufficient complementarity.
Rows~7--9 yield positive $\Delta_{\mathrm{gain}}$ yet reach only 79.6--83.8\% fused Top-1 (vs.\ 92.6\% in row~1), because deep DINO layers---whether paired with CLIP (rows~8--9) or with each other (row~7)---encode appearance-related structure less aligned with decodable scalp EEG than shallow spans (Sec.~\ref{sec:teacher_repr_diff}).
Together, these ablations indicate that effective fusion depends on cross-paradigm semantic and appearance-related supervision---multi-blur CLIP with shallow DINO---that better captures the hierarchical visual information embedded in EEG signals, rather than redundant or misaligned multi-teacher stacking.

\subsubsection{Fusion and Training Ablations}
Table~\ref{tab:fusion_training_ablation} ablates DTHA training objectives and fusion architectures under the default CLIP+DINOv2 recipe on THINGS-EEG.
Table~\ref{tab:fusion_training_ablation}(a)i ablates branch and fused-retrieval losses.
Disabling either branch SCE loss lowers Top-1 by 2.5~pp or 1.1~pp, indicating that semantic and appearance-related supervision must both remain active.
\emph{Fused-retrieval-only} training ($0{:}0{:}1$) performs substantially worse (88.8\%, $-3.8$~pp): learning both teachers jointly through $S_{\mathrm{fused}}$ alone deprives each branch of sufficient paradigm-specific supervision.
With all three losses enabled (default $1{:}1{:}0.25$), branch SCE separately supervises paradigm-specific visual cues through $S^{\mathrm{clip}}$ and $S^{\mathrm{dino}}$, while the fused-retrieval term learns similarity-level fusion weights that combine them complementarily.
We omit $\gamma{=}0$ because fusion weights are not learned, inconsistent with our test-time protocol.
Table~\ref{tab:fusion_training_ablation}(a)ii shows that the fused-retrieval weight $\gamma$ must remain balanced: $\gamma{=}0.1$ under-supervises fusion-weight learning ($-0.4$~pp), whereas $\gamma{=}1.0$ over-emphasizes the fused-retrieval loss relative to branch SCE, trending toward \emph{fused-retrieval-only} optimization ($-0.4$~pp); default $\gamma{=}0.25$ is most stable.
Table~\ref{tab:fusion_training_ablation}(b) ablates the EEG--image fusion architecture.
A shared encoder with similarity-level fusion is critical: dual EEG encoders incur a modest 0.6~pp drop, possibly from weakened co-training synergy on one shared EEG embedding, while dual projection heads ($-14.8$~pp) and feature-level fusion (${\sim}15\%$ Top-1) substantially impair or collapse retrieval.

\begin{table}[htbp]
  \caption{Fusion and training ablations on THINGS-EEG (CLIP\,+\,DINOv2, 10-subject mean; seed 2025).
  $\Delta$: Top-1 gap vs.\ default DTHA.}

  \label{tab:fusion_training_ablation}
  \centering
  \footnotesize
  \setlength{\tabcolsep}{2.5pt}
  \begin{tabular}{@{}l ccc@{}}
    \toprule
    \addlinespace[2pt]
    \multicolumn{4}{@{}l@{}}{\textit{(a) Training objective}} \\
    \addlinespace[2pt]
    Branch losses $(\alpha,\beta)$ and fused term $\gamma$ & Top-1 & Top-5 & $\Delta$ \\
    \midrule
    \textbf{Default} $(1{:}1{:}0.25)$ & \textbf{92.6} & \textbf{99.5} & -- \\
    CLIP branch only $(1{:}0{:}0.25)$ & 90.1 & 99.2 & $-2.5$ \\
    DINO branch only $(0{:}1{:}0.25)$ & 91.5 & 99.2 & $-1.1$ \\
    Fused-retrieval only $(0{:}0{:}1)$ & 88.8 & 98.9 & $-3.8$ \\
    \midrule
    Loss weight $\gamma$ ($\alpha{=}\beta{=}1$) & Top-1 & Top-5 & $\Delta$ \\
    \midrule
    $\gamma{=}0.1$ & 92.2 & 99.5 & $-0.4$ \\
    \textbf{Default} $\gamma{=}0.25$ & \textbf{92.6} & \textbf{99.5} & -- \\
    $\gamma{=}0.5$ & 92.5 & 99.5 & $-0.1$ \\
    $\gamma{=}1.0$ & 92.2 & 99.6 & $-0.4$ \\
    \midrule
    \addlinespace[2pt]
    \multicolumn{4}{@{}l@{}}{\textit{(b) EEG--image fusion architecture}} \\
    \addlinespace[1pt]
    Variant & Top-1 & Top-5 & $\Delta$ \\
    \midrule
    \textbf{Shared encoder + dual similarity fusion (default)} & \textbf{92.6} & \textbf{99.5} & -- \\
    Dual EEG encoders (separate $z_{\mathrm{CLIP}}$, $z_{\mathrm{DINO}}$) & 92.0 & 99.4 & $-0.6$ \\
    Dual projection heads (shared encoder) & 77.8 & 96.4 & $-14.8$ \\
    Feature-level fusion, scalar $w$ & 15.2 & 43.4 & $-77.4$ \\
    Feature-level fusion, per-dim.\ $w$ & 14.4 & 42.6 & $-78.1$ \\
    \bottomrule
  \end{tabular}
\end{table}

\subsection{Brain Encoder (AMSPBE) Analysis}
\label{sec:encoder_analysis}

We validate the effectiveness of AMSPBE by comparing it with established EEG encoders, ablating its core components, and analyzing the learned tri-region temporal attention curves and scalp-pattern topographies.

\subsubsection{Encoder Comparison}
As shown in Table~\ref{tab:encoder_compare}, our proposed encoder consistently outperforms both classic and recent methods on THINGS-EEG and THINGS-MEG.
On THINGS-EEG, paired one-sided Wilcoxon signed-rank tests on per-subject Top-1 ($n{=}10$), with Holm correction, confirm that AMSPBE significantly outperforms every alternative ($^{*}p{<}0.05$ for SpatialConv; $^{**}p{<}0.01$ for all others).

\begin{table}[!tbp]
  \caption{Brain-encoder comparison under a fixed DTHA recipe (seed 2025).
  EEG Top-1 stars: one-sided paired Wilcoxon vs.\ AMSPBE.}
  \label{tab:encoder_compare}
  \centering
  \footnotesize
  \setlength{\tabcolsep}{7.5pt}
  \begin{tabular}{lcccc}
    \toprule
    \multirow{2}{*}{Brain encoder} & \multicolumn{2}{c}{THINGS-EEG} & \multicolumn{2}{c}{THINGS-MEG} \\
    \cmidrule(lr){2-3}\cmidrule(lr){4-5}
    & Top-1 & Top-5 & Top-1 & Top-5 \\
    \midrule
    \textbf{AMSPBE (proposed)} &
      \textbf{92.6} & \textbf{99.5} &
      \textbf{64.6} & \textbf{87.6} \\
    \tblencoder{SpatialConv}{liu2026visualeeg}{LVBP} &
      91.5$^{*}$  & 99.5 &
      63.0 & 86.5 \\
    \tblencoder{EEGNet}{lawhern2018eegnet}{JNE'18} &
      72.1$^{**}$ & 94.0 &
      36.2 & 64.1 \\
    \tblencoder{TSConv}{song2024decoding}{NICE} &
      66.4$^{**}$ & 91.0 &
      29.5 & 60.2 \\
    \tblencoder{STAE}{wu2026ats}{ATS} &
      62.2$^{**}$ & 90.0 &
      43.3 & 72.9 \\
    \tblencoder{ShallowNet}{schirrmeister2017deep}{HBM'17} &
      59.5$^{**}$ & 87.8 &
      24.4 & 48.8 \\
    \tblencoder{DeepConvNet}{schirrmeister2017deep}{HBM'17} &
      54.5$^{**}$ & 85.2 &
      27.4 & 55.2 \\
    \tblencoder{EEGProject}{wu2025ubp}{UBP} &
      48.9$^{**}$ & 81.2 &
      29.0 & 55.6 \\
    \bottomrule
  \end{tabular}
\end{table}

\subsubsection{Component Ablation}
Table~\ref{tab:amfbe_component_ablation} ablates AMSPBE components.
Ablating either main component still leaves a functional encoder with substantial retrieval performance---70.8\% Top-1 without the $K{=}16$ scalp-pattern filter bank, and 92.1\% without TSAM and depthwise temporal convolution---while the full model reaches 92.6\%, showing that both components contribute and work best together.
Under the default DINOv2 recipe, TSAM and depthwise temporal convolution yield only modest Top-1 gains (+0.6/+0.7 pp), yet their main role is physiologically interpretable tri-region temporal dynamics (Fig.~\ref{fig:spectral_layer}(A--B)).
Filter-bank size $K$ and depthwise-convolution settings ($k$, $t_{\mathrm{pool}}$) were selected by grid search under the default DTHA recipe (Fig.~\ref{fig:eeg_encoder}).

\begin{table}[htbp]
  \caption{AMSPBE component ablation under the fixed DTHA dual-teacher recipe (fused Top-1/Top-5, \%; seed~2025).
  EEG: THINGS-EEG ($n{=}10$); MEG: THINGS-MEG ($n{=}4$).
  TSAM abbreviation is defined in Fig.~\ref{fig:eeg_encoder}.}
  \label{tab:amfbe_component_ablation}
  \centering
  \footnotesize
  \setlength{\tabcolsep}{2.5pt}
  \begin{tabular}{@{}l cccc@{}}
    \toprule
    \multirow{2}{*}{Variant} & \multicolumn{2}{c}{THINGS-EEG} & \multicolumn{2}{c}{THINGS-MEG} \\
    \cmidrule(lr){2-3}\cmidrule(lr){4-5}
    & Top-1 & Top-5 & Top-1 & Top-5 \\
    \midrule
    \textbf{Full AMSPBE (default)} &
      \textbf{92.6} & \textbf{99.5} &
      \textbf{64.6} & \textbf{87.6} \\
    \midrule
    \multicolumn{5}{@{}l@{}}{\textit{Main component ablations}} \\
    w/o spatial filter bank ($K{=}16$) &
      70.8 & 94.5 & 39.9 & 67.6 \\
    w/o TSAM \& depthwise temporal conv &
      92.1 & 99.4 & 63.4 & 88.0 \\
    \midrule
    \multicolumn{5}{@{}l@{}}{\textit{TSAM and depthwise temporal conv ablations}} \\
    w/o TSAM &
      91.9 & 99.4 & 64.0 & 85.6 \\
    w/o depthwise temporal conv &
      92.0 & 99.2 & 61.5 & 86.9 \\
    \bottomrule
  \end{tabular}
\end{table}

\subsubsection{Learned Temporal Attention and Scalp Patterns}
Given the interpretability role of TSAM discussed above, we visualize the learned tri-region temporal attention curves and scalp-pattern topographies in Fig.~\ref{fig:spectral_layer}(A--D).
The visualization reveals that AMSPBE automatically assigns higher weights to physiologically meaningful time points rather than arbitrary samples along the RSVP window.
On THINGS-EEG (Fig.~\ref{fig:spectral_layer}(A), 10-subject mean), the posterior partition shows the strongest peaks (e.g., ${\sim}76$~ms), with earliest emphasis near ${\sim}50$~ms after onset---consistent with the latency of the initial feedforward sweep to primary visual cortex~\cite{thorpe1996}.
Across partitions, a clear posterior-to-anterior lag matches hierarchical propagation along the occipito-frontal pathway~\cite{dicarlo2012how,cichy2014}: occipital responses rise first, followed by central and frontal contributions.
On THINGS-MEG (Fig.~\ref{fig:spectral_layer}(B), 4-subject mean), the posterior partition often exhibits a pronounced double peak, including a late component around ${\sim}605$~ms; the clearer early--late separation plausibly reflects MEG's longer stimulus duration ($500$~ms) and SOA ($1500{\pm}200$~ms), which attenuate RSVP-style temporal aliasing relative to THINGS-EEG ($100$~ms on, $100$~ms off).
Fig.~\ref{fig:spectral_layer}(C) displays the sixteen learned scalp-pattern topographies of the $K{=}16$ filter bank for a representative subject (Sub01), each corresponding to a distinct spatial filtering pattern over the electrode montage.
Fig.~\ref{fig:spectral_layer}(D) averages these topographies across ten subjects: occipital channels dominate, and beyond the occipito-parietal cluster (O/PO/P), a secondary hotspot over right prefrontal and frontopolar sites (Fp2, AF4/AF8, F4/F6, FC2/FC4) indicates that several patterns jointly leverage frontal and posterior visual channels.

Together, these qualitative results demonstrate the neuroscientific plausibility of AMSPBE through consistency with established visual processing findings; the salient channels and time points identified here may inform visual BCI montage design by indicating which channels and RSVP latencies carry the most decodable content.

\subsection{Regional and Spectral Decoding Analysis}
\label{sec:spatio_spectral}

\begin{figure*}[!t]
  \centering
  \includegraphics[width=\textwidth]{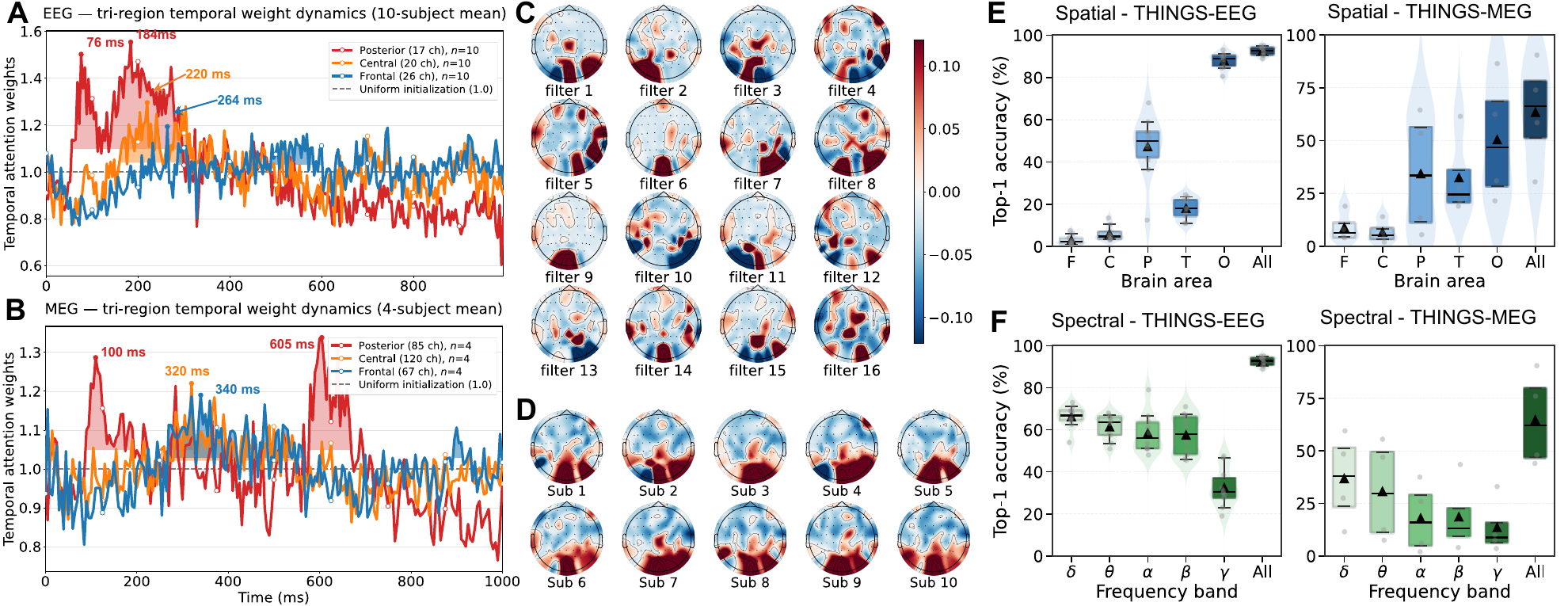}
  \caption{Spatio-spectral interpretability on THINGS-EEG and THINGS-MEG (10-subject / 4-subject mean; seed 2025).
  (A--D) Learned tri-region temporal attention and scalp-pattern topographies.
  (E) Fused Top-1 under five scalp partitions (F/C/P/T/O) vs.\ full montage (All); unselected channels zero-masked.
  (F) Fused Top-1 under $\delta$ (0.5--4), $\theta$ (4--8), $\alpha$ (8--13), $\beta$ (13--30), $\gamma$ (30--100; MEG clipped to Nyquist$-1$) vs.\ broadband (All); zero-phase 4th-order Butterworth.}
  \label{fig:spectral_layer}
\end{figure*}

Sec.~\ref{sec:encoder_analysis} characterized AMSPBE with full-montage temporal attention and scalp-pattern analyses; we next examine how decodable visual content is distributed across scalp regions and frequency bands (Fig.~\ref{fig:spectral_layer}(E--F)).

\subsubsection{Spatial and Spectral Decoding}
We first ask which scalp partitions carry decodable visual content.
Fig.~\ref{fig:spectral_layer}(E) reports fused Top-1 when training and testing on each of five standard partitions---frontal (F), central (C), parietal (P), temporal (T), and occipital (O)---following~\cite{song2025recognizing}, versus the full montage (All); unselected channels are zero-masked while preserving layout.
On THINGS-EEG, occipital channels alone reach ${\sim}88\%$ and dominate single-partition performance; parietal contributes moderately (${\sim}50\%$), and frontal, central, and temporal subsets remain comparatively low; the full montage reaches 92.6\%.
THINGS-MEG shows the same posterior emphasis (occipital ${\sim}50\%$; full sensors ${\sim}65\%$), supporting cross-modality consistency.

We then examine frequency content.
Fig.~\ref{fig:spectral_layer}(F) reports fused Top-1 under $\delta$, $\theta$, $\alpha$, $\beta$, or $\gamma$ band-pass filtering versus broadband inputs (All).
On EEG, $\delta$--$\beta$ retain substantial decoding (${\sim}58$--$68\%$), whereas $\gamma$-only inputs drop sharply (${\sim}32\%$); broadband (All) reaches 92.6\%.
MEG shows the same downward trend with frequency ($\gamma$ lowest, ${\sim}12\%$), indicating that decodable RSVP-evoked content is concentrated in lower-frequency components.

\subsubsection{Partition-Wise DTHA Decoding}
Under regional partitioning, Table~\ref{tab:regional_branch} further evaluates two additional groupings---posterior V17 (17 P/PO/O channels) vs.\ non-posterior NV46---beyond the five standard partitions in Fig.~\ref{fig:spectral_layer}(E), reporting per-branch and fused DTHA retrieval.
Decoding is highest in posterior regions (occipital and V17), where the DINO branch clearly outperforms CLIP (e.g., V17: $87.1\%$ vs.\ $78.4\%$; fused $90.9\%$), consistent with appearance-related dominance in these channels.
On the full montage, fused retrieval reaches $92.6\%$ and the DINO branch leads CLIP by $7.8$~pp ($89.0\%$ vs.\ $81.3\%$), in line with this posterior emphasis.
Outside occipital/V17, performance falls sharply, yet NV46 still retains $31.8\%$ fused Top-1---far above the 200-way chance level---indicating complementary non-posterior contributions.
Learned fusion weights mirror this trend ($w_{\mathrm{dino}}{\approx}0.5$ in occipital/V17 vs.\ ${\approx}0.32$ elsewhere), with a relative shift toward the semantic CLIP branch outside these regions---consistent with our mechanism--paradigm correspondence view.

\begin{table}[!tbp]
  \caption{Partition-wise DTHA retrieval under regional partitioning (THINGS-EEG, 10-subject mean; seed~2025).
  Train/test on masked channels only; unselected channels are zero-masked.
  CLIP and DINO: branch Top-1 (\%); Fused: fused Top-1 (\%); DINO$-$CLIP: branch gap (pp).
  $w_{\mathrm{dino}}$: subject-averaged learned fusion weight on the DINO branch ($w_{\mathrm{clip}}{+}w_{\mathrm{dino}}{=}1$).}
  \label{tab:regional_branch}
  \centering
  \footnotesize
  \setlength{\tabcolsep}{3.7pt}
  \begin{tabular}{lccccc}
    \toprule
    Region (channels) & CLIP & DINO & Fused & DINO$-$CLIP & $w_{\mathrm{dino}}$ \\
    \midrule
    Occipital (8)  & 73.5 & 85.3 & 87.9 & $+11.8$ & 0.502 \\
    Parietal (16)  & 40.6 & 40.2 & 47.3 & $-0.4$  & 0.366 \\
    Temporal (10)  & 16.3 & 13.0 & 18.0 & $-3.4$  & 0.323 \\
    Central (14)   & 5.9  & 4.4  & 6.0  & $-1.6$  & 0.308 \\
    Frontal (15)   & 2.9  & 1.7  & 3.0  & $-1.2$  & 0.326 \\
    \midrule
    Posterior V17 (17)      & 78.4 & 87.1 & 90.9 & $+8.7$ & 0.492 \\
    Non-posterior NV46 (46) & 28.3 & 24.6 & 31.8 & $-3.7$ & 0.324 \\
    \midrule
    Full montage (63)       & 81.3 & 89.0 & 92.6 & $+7.8$ & 0.489 \\
    \bottomrule
  \end{tabular}
\end{table}

\begin{figure*}[!t]
  \centering
  \includegraphics[width=\textwidth]{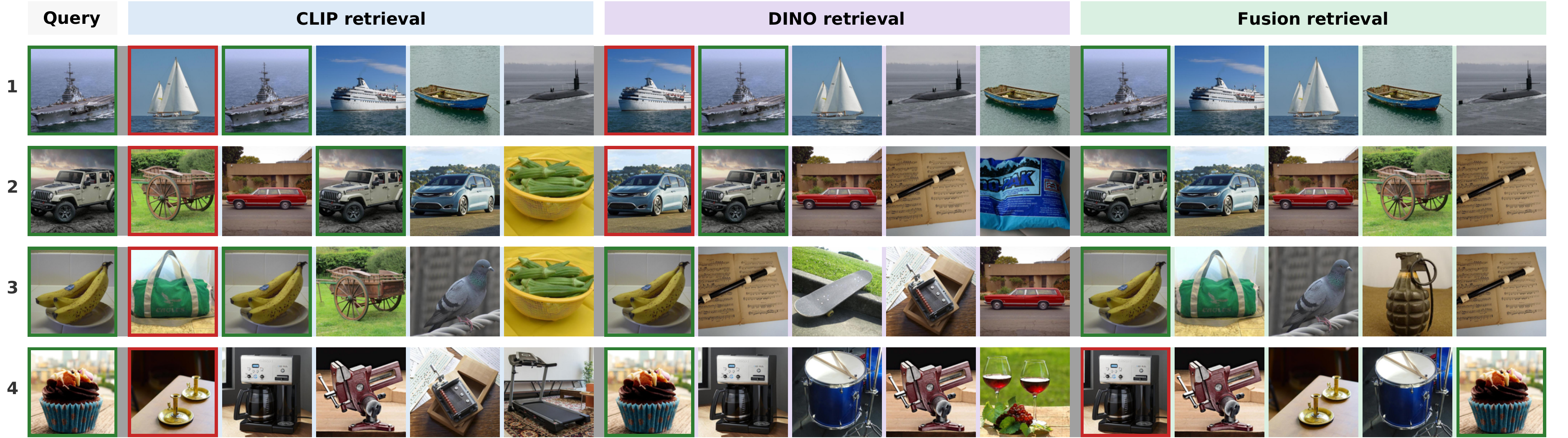}
  \caption{CLIP-branch, DINO-branch, and fused Top-1 retrieval on four representative THINGS-EEG test queries.
  Green/red borders mark correct/incorrect Top-1 hits.
  Rows~1--2 (aircraft carrier, SUV): fusion-gain cases---both branches miss and fusion succeeds; row~3 (banana): CLIP fails but DINO succeeds and fusion follows; row~4 (cupcake): DINO succeeds but fusion fails---an occasional conflict case under similarity-level fusion.}
  \label{fig:retrieval_cases}
\end{figure*}

\begin{figure}[htbp]
  \centering
  \includegraphics[width=\columnwidth]{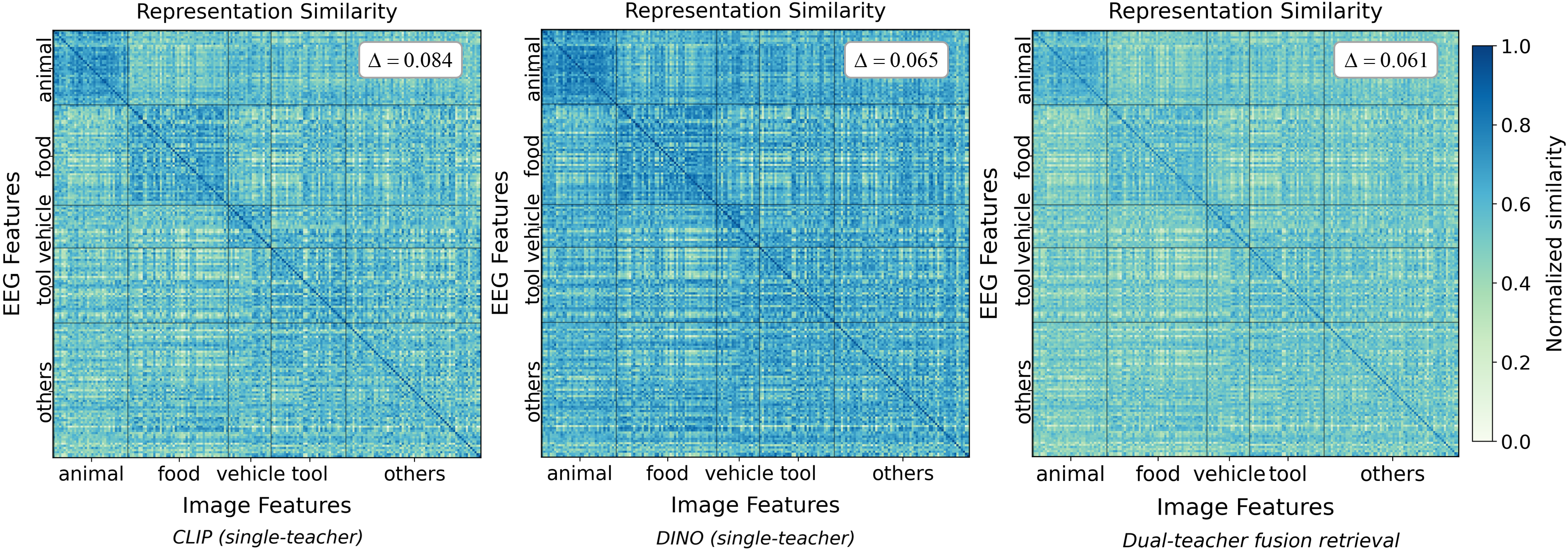}
  \caption{Cross-modal EEG--image similarity on THINGS-EEG. $\bar{s}_{\mathrm{intra}}$ and $\bar{s}_{\mathrm{inter}}$ denote mean normalized similarities within and across superordinate blocks; $\Delta=\bar{s}_{\mathrm{intra}}-\bar{s}_{\mathrm{inter}}$ (larger $\Delta$ = stronger intra-category aggregation).}
  \label{fig:brain_image_rsa}
\end{figure}

\begin{figure}[htbp]
  \centering
  \includegraphics[width=\columnwidth]{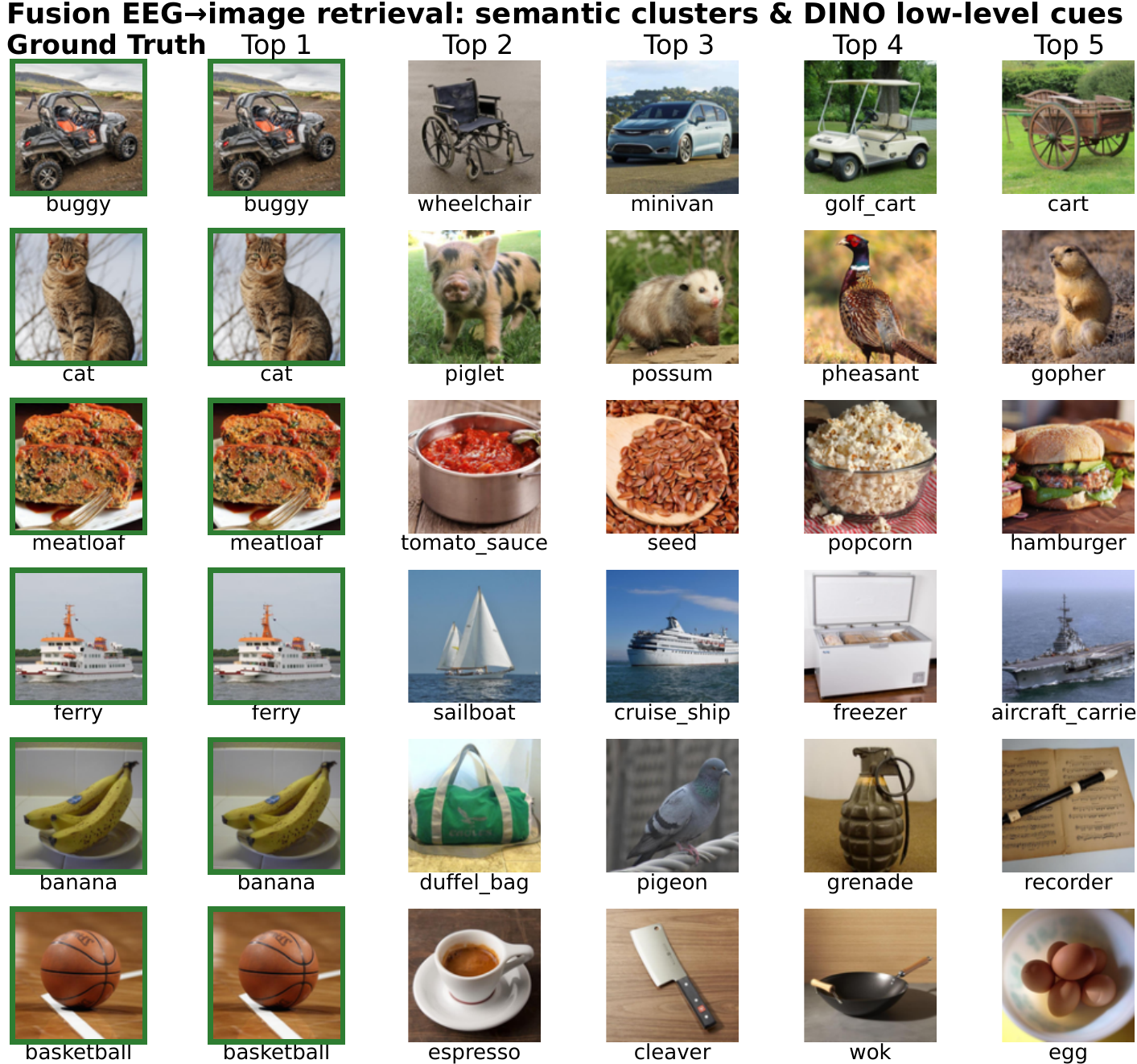}
  \caption{Representative fused Top-5 EEG-to-image retrieval on THINGS-EEG test queries.
  Each row shows the query and top-5 gallery hits ranked by $S_{\mathrm{fused}}$ (green border: query and correct Top-1 hit).
  All shown queries recover the correct target at Top-1; ranks 2--5 illustrate how semantic and appearance-related cues combine in the ranked gallery.
  Buggy, cat, and meatloaf: remaining hits are largely superordinate-consistent neighbors; banana and basketball: remaining hits are largely appearance-related matches across labels; for \textit{ferry}, remaining hits mix ship-consistent matches (e.g., sailboat, cruise ship) with the appearance-driven hit \textit{freezer}.}
  \label{fig:fusion_retrieval_top5}
\end{figure}

\subsection{Qualitative Analysis}
\label{sec:qualitative_retrieval}

Having established DTHA's effectiveness through the quantitative results above, we further dissect the mechanisms underlying complementarity-driven DTHA through qualitative analysis on THINGS-EEG.
Following~\cite{song2024decoding}, the $200$ test concepts are grouped into five superordinate categories (animal, food, vehicle, tool, and others).
On this grouping, we perform category-structured cross-modal RSA on the test set, followed by retrieval case analysis on representative test queries.

\subsubsection{Cross-Modal Representational Similarity}
Fig.~\ref{fig:brain_image_rsa} shows cross-modal representational similarity matrices (RSMs) on the THINGS-EEG test set (10-subject mean) under CLIP and DINO single-teacher alignment, alongside DTHA fused retrieval.
CLIP single-teacher alignment exhibits strong intra-category aggregation (bright diagonal blocks; $\Delta{=}0.084$), confirming category-level semantic structure, together with nuanced off-diagonal inter-class similarity as expected under multi-blur supervision.
DINO single-teacher alignment shows a weaker category diagonal ($\Delta{=}0.065$) but notable off-diagonal inter-class similarity, indicating appearance-related structure beyond superordinate category labels.
DTHA fused retrieval yields a softer category diagonal ($\Delta{=}0.061$) yet the strongest Top-1 accuracy (Table~\ref{tab:things_eeg_results}), indicating that retrieval accuracy is not explained by sharper within-category similarity alone, but by a hybrid of category-level semantics and appearance-related visual cues.

\subsubsection{Retrieval Case Analysis}
To provide an intuitive illustration of fused retrieval, we analyze representative retrieval cases on test queries.
As shown in Fig.~\ref{fig:fusion_retrieval_top5}, fused retrieval recovers the correct target at Top-1, while the remaining Top-5 slots mix superordinate-consistent and appearance-related neighbors.
Together, these results indicate that retrieval engages both semantic and appearance-related cues.
Fig.~\ref{fig:retrieval_cases} contrasts CLIP-branch, DINO-branch, and fused Top-1 retrieval on four representative queries.
Rows~1--2 (aircraft carrier, SUV): both branches miss the target while fusion recovers the correct image---clear fusion-gain cases when neither teacher alone succeeds.
Row~3 (banana): the CLIP branch fails but the DINO branch retrieves correctly; fusion follows the successful branch and also hits the target.
Row~4 (cupcake): the DINO branch succeeds yet fusion fails when branch rankings conflict---the fused score favors a CLIP-branch semantic distractor over the DINO-correct target, illustrating an occasional conflict case under similarity-level fusion despite overall complementary gains (Table~\ref{tab:things_eeg_results}).
Collectively, cross-paradigm dual-teacher supervision provides complementary visual cues, supporting more complete extraction of hierarchical visual information from EEG and thereby improving retrieval accuracy.

\section{Discussion and Conclusion}
\label{sec:conclusion}

Motivated by a \textbf{mechanism--paradigm correspondence} between hierarchical visual processing and VFMs trained under distinct paradigms, we proposed \textbf{DTHA}, a cross-paradigm complementarity-driven dual-teacher framework that aligns EEG to complementary semantic (multi-blur CLIP) and appearance-related (shallow DINOv2) teachers through dual-branch supervision and similarity-level fused retrieval.
We further introduced \textbf{AMSPBE}, which pairs strong decoding with interpretable tri-region temporal dynamics and scalp-pattern topographies.

On THINGS-EEG zero-shot EEG-to-image retrieval, DTHA achieves 92.4\% Top-1 accuracy, surpassing the previous state-of-the-art by 12.4 percentage points.
Teacher-representation analyses, pairing ablations, and fusion-training studies showed that fusion gain arises from cross-paradigm complementarity rather than arbitrary multi-teacher stacking, while qualitative cases indicated hybrid decoding that jointly reflects semantic and appearance-related cues.
Spatial, spectral, and partition-wise analyses, together with THINGS-MEG evaluation, further supported biological plausibility and cross-modality generalization.

One limitation of this work is that inter-subject retrieval remains far below intra-subject performance.
Most existing EEG visual decoding methods have not specifically targeted the cross-subject generalization challenge, which remains an important direction for our future work.

In summary, this work shows that complementary supervision from cross-paradigm VFMs enables more complete extraction of hierarchical visual information from brain activity, improving both zero-shot retrieval accuracy and interpretability of EEG visual decoding.

\balance
\bibliographystyle{IEEEtran}
\bibliography{refs}

\end{document}